\documentclass[runningheads]{llncs}
\usepackage[textsize=small]{todonotes}

\usepackage[T1]{fontenc}
\usepackage{graphicx}
\usepackage{booktabs}
\usepackage[misc]{ifsym}
\newcommand{\corr}{(\Letter)}
\usepackage{amsmath}
\usepackage{amssymb}
\usepackage{mwe}
\usepackage{hyperref}
\usepackage{blindtext}
\usepackage{amsfonts}
\usepackage{algorithm}
\usepackage{algpseudocode}
\usepackage{float}
\usepackage{tabularx}
\usepackage{xcolor}
\usepackage{subcaption}
\usepackage{makecell}
\usepackage{color}
\usetikzlibrary{calc}
\usepackage[title]{appendix}
\usepackage{xr}

\usepackage{fontawesome5}
\usepackage{subcaption}
\usepackage{array}

\usepackage{environ}
\NewEnviron{myhideenv}{%
  \setbox0\vbox{%
   \let\xxwrite\write
   \protected\def\write{\immediate\xxwrite}%
   {\tiny XX\BODY XX}}
  }

\newcommand*{\mybox}[2]{\colorbox{#1!30}{\parbox{.88\linewidth}{#2}}}
\newtheorem{observation}{Observation}

\usepackage{soul}
\sethlcolor{lightgray}

\begin{document}

\title{PSI-PFL: Population Stability Index for Client Selection in non-IID Personalized Federated Learning}

\titlerunning{PSI-PFL: PSI for Client Selection in non-IID PFL}


\author{
Daniel M. Jimenez-Gutierrez \corr \orcidID{0000-0002-0305-814X} \and David Solans \corr \orcidID{0000-0001-6979-9330} \and Mohammed Elbamby \orcidID{0000-0003-0078-3933} \and Nicolas Kourtellis \orcidID{0000-0002-5674-1698}}





\institute{Sapienza University of Rome, Rome, Italy \email{danielmauricio.jimenezgutierrez@uniroma1.it}
\and
Telefonica, Barcelona, Spain \email{david.solansnoguero@telefonica.com}
}

\maketitle              

\begin{abstract}

Federated Learning (FL) enables decentralized machine learning (ML) model training while preserving data privacy by keeping data localized across clients. However, non-independent and identically distributed (non-IID) data across clients poses a significant challenge, leading to skewed model updates and performance degradation. Addressing this, we propose PSI-PFL, a novel client selection framework for Personalized Federated Learning (PFL) that leverages the Population Stability Index (PSI) to quantify and mitigate data heterogeneity (so-called non-IIDness). Our approach selects more homogeneous clients based on PSI, reducing the impact of label skew, one of the most detrimental factors in FL performance. Experimental results over multiple data modalities (tabular, image, text) demonstrate that PSI-PFL significantly improves global model accuracy, outperforming state-of-the-art baselines by up to 10\% under non-IID scenarios while ensuring fairer local performance. PSI-PFL enhances FL performance and offers practical benefits in applications where data privacy and heterogeneity are critical. 
The code for our solution will be released after receiving the paper's acceptance.

\end{abstract}


\section{Introduction}
\label{sec:intro}

Federated Learning (FL)~\cite{mcmahan2017communication} has emerged as a transformative paradigm in machine learning (ML), enabling decentralized training across multiple clients while preserving data privacy. Unlike centralized learning (CL), where all data is aggregated on a central server, FL operates on distributed datasets, often characterized by non-IID (non-independent and identically distributed) distributions. This so-called non-IIDness (also known as data heterogeneity, non-IID data), arising from client-specific behaviors and contexts, presents significant challenges, including skewed model updates and degraded global performance~\cite{g2024noniiddatafederatedlearning}.

Personalized FL (PFL) extends FL by tailoring models to individual clients, allowing for better adaptation to heterogeneous data. Tan et al.~\cite{tan2022towards} classify PFL approaches into regularization-based methods, which adjust local objectives to align with global goals, and selection-based strategies that prioritize specific clients for training. The solution proposed in this work falls into the \emph{selection-based} category. 

This paper leverages the Population Stability Index (PSI), a metric widely used in credit scoring to measure distributional shifts~\cite{yurdakul2018statistical,du2023proposed}, as a novel tool for client selection in FL. The PSI quantifies the degree of heterogeneity among clients, enabling the identification of more homogeneous subsets for training. We demonstrate the application of the PSI to address label skew, as this form of skewness poses the most significant challenge to model performance in FL~\cite{g2024noniiddatafederatedlearning}. Extending the use of PSI to other types of data skewness is left as an avenue for future exploration. 

\paragraph{\textbf{Motivation}}
Non-IID data, where data distributions differ across examples or individuals, presents a fundamental challenge in many ML systems, including FL.
FL relies on decentralized data across clients, often reflecting varying distributions due to diverse user behaviors, preferences, or contexts. This heterogeneity can lead to skewed model updates, slower convergence rates, and a consequential degraded global model performance. Diagnosing and addressing non-IID-ness 
is critical for ensuring that FL systems can achieve fairness, robustness, and accuracy across various applications.

Additionally, heterogeneity diagnostics in FL 
remains an open challenge, as highlighted by Pei et al.~\cite{pei2024review} and Li et al.~\cite{li2020federated}, who outline key research objectives in this area. These include developing simple diagnostics to quickly assess the level of heterogeneity in federated networks a priori, exploring methods for conducting heterogeneity analysis and diagnosis while preserving privacy, establishing effective ways to quantify heterogeneity, and investigating whether current or novel definitions of heterogeneity can be leveraged to design federated optimization methods that enhance convergence empirically and theoretically.

\paragraph{\textbf{Contribution.}}
The following points summarize the contributions of our paper:

\begin{enumerate}  
    \item We use the PSI as a metric to efficiently and accurately quantify the degree of non-IIDness in an FL system, highlighting its advantages and comparing it with existing non-IID metrics.  
    \item We propose \textbf{PSI-PFL: Population Stability Index for Personalized Federated Learning}, a novel approach to mitigate performance degradation in FL models as non-IIDness increases, leveraged on the PSI-based client selection.  
    \item Our PSI-PFL strategy improves FL global model accuracy by up to \textbf{10\%} compared to state-of-the-art baselines under non-IID scenarios.
\end{enumerate}

To the best of our knowledge, this is the first work to employ PSI in FL to quantify data heterogeneity and select clients to enhance global model performance.

\paragraph{\textbf{Paper structure.}}

In the next section, we provide the background and related literature. Section~\ref{sec:strategy} explains our proposed PSI-PFL approach in detail. In Section~\ref{sec:experiments}, we present the experiments carried out and the results obtained. Finally, we conclude and provide some limitations and future work in Section~\ref{sec:conclusion}.

\section{Background and Related Work}
\label{sec:related}
\paragraph{\textbf{Federated Learning Foundations.}}
\label{sec:problem}

\begin{figure}[ht]
\centering
\includegraphics[width=0.99\textwidth]{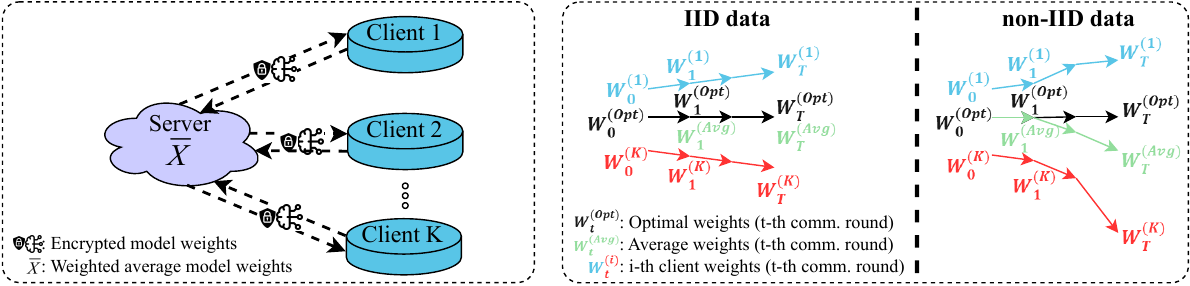}
\caption{(Left) FL usual training process with FedAvg. (Right) Weight divergence illustration for FL with non-IID data}
\label{fig:fl-diagram}
\end{figure}

In Fig.~\ref{fig:fl-diagram} (left), we illustrate the usual training process of a model in a FL setting. It begins with model initialization, where a central server shares an initial global model with participating $K$ clients. Each client performs local computation by training the model on its private dataset, generating updates (a.k.a. weights) that encapsulate learned patterns without exposing sensitive information. These updates are transmitted back to the server, often encrypted using privacy-preserving techniques such as Secure Multi-Party Computation (MPC)~\cite{bohler2021secure} or differential privacy (DP)~\cite{apple_privacy_2017} to enhance security. The server then performs model aggregation, combining the local updates (e.g., via Federated Averaging - FedAvg) to refine the global model, which is redistributed to clients for further iterations. This iterative process continues until the model converges.

A significant challenge in FL arises from the presence of non-IID data across participating clients~\cite{g2024noniiddatafederatedlearning}. Unlike IID data, where local datasets share similar distributions, non-IID data introduces heterogeneity in the data distribution across clients, as illustrated in Fig.~\ref{fig:fl-diagram} (right). This heterogeneity can lead to divergence between the local model weights ($W_t^{i}$) and the global optimal weights ($W_t^{(Opt)}$) during training. While IID data allows local updates to align closely with the global objective, non-IID data causes local updates to deviate significantly, resulting in slower convergence and suboptimal global models. Additionally, as illustrated in the figure, ($W_t^{(Avg)}$) can fall into local optima, failing to capture the global optimum and consequently hindering the model's performance. 

Measuring and tackling significant disparities in label distribution among clients participating in an FL model is a well-recognized challenge that impacts both the model's performance and its convergence during training~\cite{aggregators_FL,hsieh2020non}. To quantify the degree of non-IIDness, various metrics have been proposed~\cite{g2024noniiddatafederatedlearning}. In the context of PFL, two prominent strategies for mitigating these challenges are regularization-based and selection-based approaches~\cite{tan2022towards}. These topics are thoroughly analyzed in the sections that follow.

\paragraph{\textbf{Non-IIDness quantification in FL.}}
The Hellinger distance (HD) is a widely utilized metric for assessing differences in client data distributions. For instance, Tan et al.~\cite{tan2023privacy} employed HD to measure the gap between clients' local data distributions and an ideal balanced distribution, aiming to reduce model divergence. Similarly, Jimenez et al.~\cite{jimenez2024fedartml} extensively analyzed HD's performance compared to other metrics in FL scenarios, concluding that HD offers greater granularity and is better suited for cross-device FL settings.

The Jensen-Shannon distance (JSD) is an alternative metric for measuring non-IIDness in FL. Ahmed et al.~\cite{ahmed2023semisupervised} employed JSD as a top-level comparison metric, analyzing divergence matrices from attention heads to distinguish between the base and news-trained models in their semi-supervised FL framework. To mitigate label distribution skew, Xu et al.~\cite{xu2024fblg} leveraged JSD to construct a local graph based on clients' local losses, enabling the selection of similar clients for aggregation and promoting a more consistent global model.

The Earth mover's distance (EMD), also known as Wasserstein distance, is another widely used metric to quantify differences among client distributions. Zhang et al.~\cite{chen2022emd} proposed ACSFed, which uses EMD and prior training performance to dynamically select clients for each communication round. Clients with greater statistical heterogeneity or poor training performance are prioritized for later training phases. Zhao et al.~\cite{zhao2023non} employed EMD to quantify the level of non-IID difference between the label distribution of the clients and the central distribution.

\paragraph{\textbf{Regularization-based solutions for non-IID data.}}
FedProx, introduced by Li et al.~\cite{li2020federated}, tackles the challenges of heterogeneity in FL by extending and reparameterizing the FedAvg algorithm. This framework begins by initializing global model parameters, followed by computing local loss functions, applying proximal gradient descent for local updates, and aggregating these updates to refine the global model. A key innovation in FedProx is including a regularization term in the local objective function, which aims to address data heterogeneity.

FedAdagrad, FedYogi, and FedAdam are adaptive federated optimization methods designed to address client heterogeneity challenges in FL~\cite{reddi2020adaptive}. These methods extend the FedOpt framework by using adaptive optimizers on the server side while employing SGD on clients, ensuring compatibility with cross-device FL. FedAdagrad leverages per-coordinate learning rates accumulated over time, making it effective for sparse gradients. FedYogi modifies the second-moment estimation to ensure stability in non-convex settings, addressing issues like heavy-tailed gradient distributions. FedAdam combines first- and second-moment estimates with exponential decay, offering robustness across diverse tasks. These methods reduce sensitivity to hyperparameter tuning compared to traditional approaches like FedAvg, as demonstrated through theoretical analysis and extensive empirical evaluations.

FedAvgM~\cite{hsu2019measuring} is an enhancement to the FedAvg algorithm designed to address challenges posed by non-IID data in FL. Unlike vanilla FedAvg, which aggregates client updates directly, FedAvgM incorporates server momentum by maintaining a running accumulation of gradient updates. This modification dampens oscillations and improves convergence stability, particularly under highly skewed data distributions. Experiments demonstrate that FedAvgM consistently outperforms FedAvg in classification accuracy across varying levels of data heterogeneity. For instance, in scenarios with extreme non.IIDness, FedAvgM achieves significantly higher accuracy and approaches centralized learning performance. Despite its benefits, FedAvgM introduces additional hyperparameter tuning complexity, particularly for the momentum coefficient and effective learning rate, which must be carefully adjusted to balance client update variance and convergence speed.

\paragraph{\textbf{Selection-based solutions for non-IID data.}}
To mitigate the effects of non-IID data, Cho et al.~\cite{pmlr-v151-jee-cho22a} showed that prioritizing clients with higher local loss can accelerate error convergence. Building on this idea, they proposed the Power-of-Choice (PoC) approach, which provides a flexible trade-off between convergence speed and solution bias. Their method achieves up to three times faster convergence and improves test accuracy by 10\% compared to random client selection. \emph{However, as we illustrate in this paper, client selection can be improved by leveraging PSI}, enabling more precise and efficient client selection that enhances accuracy by concentrating training efforts on the most homogeneous clients. 

Wolfrath et al.~\cite{wolfrath2022haccs} introduced HACCS, a client selection strategy that groups devices based on similar data histograms and selects the fastest devices (i.e., those with lower latency) from each cluster to participate in model training. HACCS demonstrates resilience to individual device dropouts as long as other devices in the system share similar data distributions. This approach ensures a balanced representation of diverse data distributions and accelerates model convergence by focusing on computationally efficient clients.

Li et al.~\cite{li2022fedcls} introduced FedCLS, an FL client selection algorithm that utilizes group label information to enhance efficiency. FedCLS optimizes client selection by calculating the Hamming distance between one-hot encoded labels, ensuring that selected clients contribute more effectively to the global model. Their results demonstrate that FedCLS surpasses FedAvg with random client selection regarding convergence speed, test accuracy, and overall stability. Additionally, the algorithm's ability to prioritize clients with diverse label distributions helps mitigate the impact of non-IID data, further improving model performance. This makes FedCLS particularly suitable for scenarios involving heterogeneous data environments, where maintaining accuracy and stability is critical.

\vspace{4pt}
\noindent
\textbf{Contributions.}
We differentiate from the non-IIDness quantification, regularization, and selection-based approaches mentioned by quantifying clients' distribution dissimilarities using the PSI to select homogeneous clients participating in the training process, improving the performance compared to state-of-the-art solutions.
In this work, we used all the previous approaches to tackle non-IIDness as baselines, adding the CL and FedAvg.


\section{PSI-PFL Proposed Solution}
\label{sec:strategy}

\begin{figure}[ht]
\centering
\includegraphics[width=0.99\textwidth]{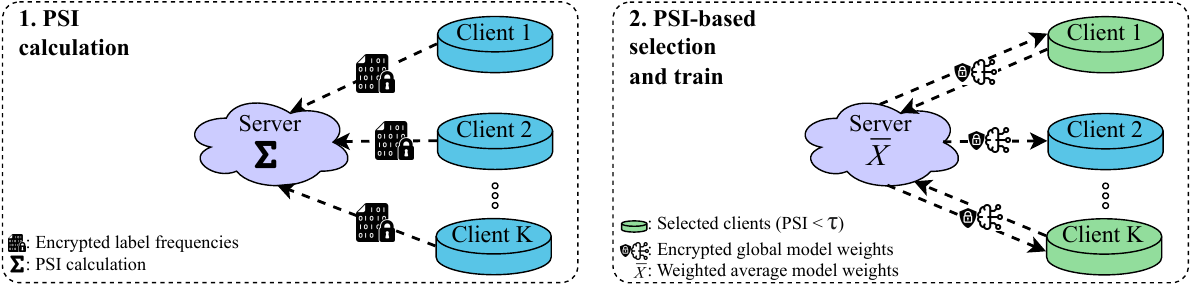}
\caption{High-level PSI-PFL training process.}
\label{fig:high-level-psi-pfl}
\end{figure}

This section explains our PSI-PFL solution, with its high-level training process illustrated in Fig.~\ref{fig:high-level-psi-pfl}. During this process, each client shares the frequency of its labels with the server. The server then computes the PSI based on the aggregated label distribution, which is derived by summing the label frequencies from all clients. Subsequently, clients with a PSI value below a predefined $\tau$ threshold are selected for model training. Once the global model has converged, it is distributed to all clients, including those not actively participating in the training phase.

In the following subsections, we explain the details of the calculation of the PSI for the clients participating in the FL training process.

\subsection{PSI calculation}
Considering the label skew case, the formula to calculate the $PSI$~\cite{siddiqi2012credit} for a client $i$ is presented in Eq.~\ref{eq:psi_formula}. Notice that the same formulation has appeared in the statistical literature as "J divergence"~\cite{lin1991divergence}:

\begin{equation} \label{eq:psi_formula}
\begin{aligned} 
     PSI_i^L = \sum_{c=1}^{C}{\left(P(y=c) - P_i(y=c)\right)\ln\left(\frac{P(y=c)}{P_i(y=c)}\right)}
\end{aligned}
\end{equation}

\noindent where $P(y=c)$ represents the global probability mass function (pmf) of the label $y$ for class $c$, which is calculated by the server with the (aggregated) label frequencies received from the clients; $P_i(y=c)$ is the client $i$ pmf of the label $y$ for class $c$, and $C$ is the maximum number of classes of $y$. Notice that the superscript $L$ represents the calculations for the label skew case.


  

Given that $PSI$ is calculated for each client, the non-IIDness quantification of the FL system in the label skew case can be determined by computing the weighted average, denoted as $WPSI^L$. Specifically, this is expressed as in Eq.~\ref{eq:wpsi_formula}:

\begin{equation} \label{eq:wpsi_formula}
\begin{aligned} 
     WPSI^L = \sum_{i=1}^{K}{\frac{n_i}{N}PSI_i^L}
\end{aligned}
\end{equation}

where $K$ is the total number of clients, $n_i$ is the number of examples in the $i$th client, and $N$ is the total number of examples across all clients. The superscript $L$ indicates that the calculation pertains to the label skew case. Notice that low values of $PSI_i$ and $WPSI^L$ suggest a more homogeneous environment, whereas higher values indicate a greater degree of non-IID characteristics.

\subsection{PSI-based client selection and train}
After calculating the PSI for each client on the server, selecting the devices that will actively participate in the global model estimation is necessary. Clients with a PSI below a predefined $\tau$ threshold are chosen for this process. The rationale behind this selection is that PSI measures the degree of non-IIDness among clients, with lower PSI values indicating less heterogeneity. By prioritizing more homogeneous clients, the global model can be trained to better represent all clients while excluding those with high heterogeneity that could negatively impact model convergence.

The $\tau$ threshold is a crucial hyperparameter in PSI-PFL that requires careful calibration based on the specific characteristics of the participating clients. We propose the following systematic approach to determine the optimal $\tau$:

\begin{enumerate}
\item Compute the $PSI$ for each client using Eq.~\ref{eq:psi_formula}.
\item Analyze the distribution of $PSI$ values across all clients. We select candidate $\tau$ thresholds from this distribution's 10th, 25th, 50th, 75th, and 90th percentiles. This ensures a comprehensive evaluation across diverse threshold values.
\item For each candidate $\tau$, train the FL model and evaluate its performance. Select the optimal $\tau$ threshold based on the highest value of our chosen performance metric (e.g., global accuracy).
\end{enumerate}

This methodology allows for a data-driven, systematic approach to threshold selection, ensuring that the chosen $\tau$ is well-suited to the specific characteristics of the client population. The efficacy of this approach is empirically validated in Section~\ref{sec:experiments}.

\subsection{PSI vs. alternative non-IID metrics}

The $WPSI^L$ stands out among metrics for quantifying non-IIDness in FL, offering distinct advantages over alternatives like Hellinger (HD), Jensen-Shannon (JSD), and Earth Mover's (EMD) distances~\cite{jimenez2024fedartml}. These specific metrics were selected as competitors for the $WPSI^L$ due to their established use in comparing probability distributions and their relevance in measuring data divergence in ML. HD and JSD are particularly noteworthy for their ability to quantify similarities between distributions~\cite{goussakov2020hellinger}, while EMD provides a measure of the "work" required to transform one distribution into another~\cite{rubner2000earth}, making them suitable benchmarks for assessing non-IID data in FL scenarios. Unlike these metrics, $WPSI^L$ provides granular client-level insights, efficiently identifying diverse and non-diverse clients with minimal computational overhead. This unique combination of detailed analysis and computational efficiency makes PSI an important tool for analyzing non-IID data in FL settings.

\begin{figure}[ht]
\centering
\includegraphics[width=0.9\textwidth]{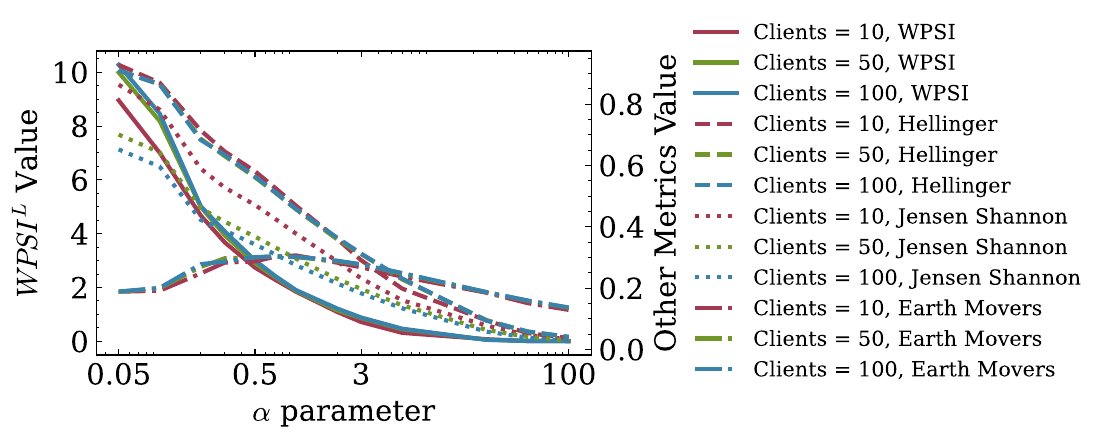}
\caption{Relationship between $WPSI^L$ and non-IIDness for the Dirichlet partition protocol.}
\label{fig:wpsi_noniidness_dirichlet_all_CLI_LS_all_Alp}
\end{figure}

To empirically validate the PSI's efficacy, we conducted a comprehensive experiment including diverse datasets, non-IIDness levels and partition protocols, random seeds, and client numbers (see Section~\ref{sec:experiments}). We partitioned the centralized datasets using the Dirichlet~\cite{jimenez2024fedartml} protocol and subsequently calculated the $WPSI^L$, HD, JSD, and EMD metrics. Fig.~\ref{fig:wpsi_noniidness_dirichlet_all_CLI_LS_all_Alp} illustrates the relationship between these metrics and the degree of non-IIDness, represented by $\alpha$. Our analysis revealed several noteworthy findings. First, the $WPSI^L$ metric exhibits an exponential decay relationship with increasing non-IIDness for both partition protocols. In contrast, JSD and HD demonstrate a more linear relationship with non-IIDness across both protocols. Interestingly, the EMD metric proved less reliable for quantifying non-IIDness due to its non-monotonic behavior, resulting in scenarios where different levels of non-IIDness yield identical metric values. These observations highlight the varying effectiveness of different metrics in capturing the nuances of data distribution in non-IID scenarios, with $WPSI^L$ potentially offering a more discriminative measure for non-IIDness compared to JSD and HD.


\begin{figure}[ht]
\centering
\includegraphics[width=0.7\textwidth]{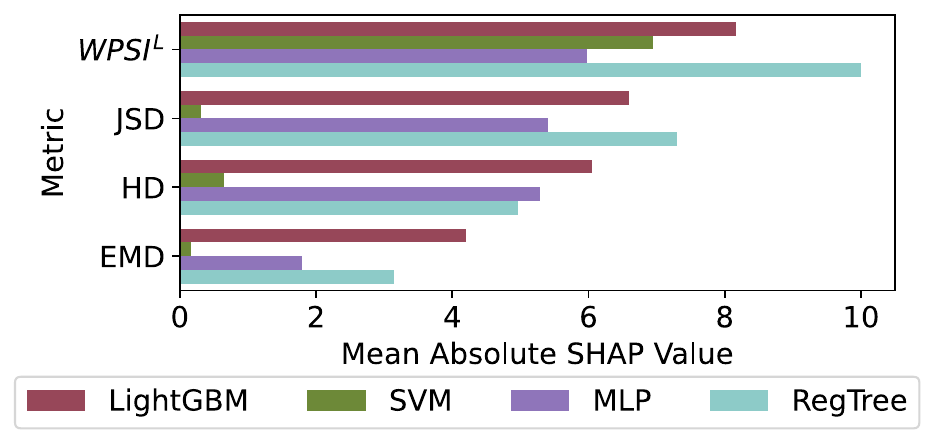}
\caption{Feature importance of models predicting non-IIDness for the Dirichlet partition protocol.}
\label{fig:wpsi_feat_imp_all_CLI_LS_dirichlet_Alp}
\end{figure}

Subsequently, we employed diverse ML models: LightGBM, Support Vector Machine (SVM), Multilayer Perceptron (MLP), and Regression Tree (RegTree). These models were trained using the calculated metrics as features to predict the $\alpha$ value for the Dirichlet partition protocol. We then analyzed the feature importance derived from each trained model to determine which metrics were most effective in quantifying and predicting non-IIDness, depicted in Fig.~\ref{fig:wpsi_feat_imp_all_CLI_LS_dirichlet_Alp}. Across all models, this analysis consistently identified $WPSI^L$ as the most significant predictor of non-IIDness. This robust performance across different ML algorithms further supports the conclusion that $WPSI^L$ is a superior metric to the others evaluated in this study for characterizing non-IID data distributions in federated settings.

\mybox{gray}{\observation{$WPSI^L$ is consistently the most significant predictor for the data heterogeneity parameter (non-IIDness).}}

\section{Experiments and Results}
\label{sec:experiments}
In this section, we first outline the simulation setup to ensure the reproducibility of our work. Then, we present the key results obtained from the analysis. 
We examine the following three experimental questions (EQs) that address key aspects of PSI-PFL and its comparison against state-of-the-art baselines:

\begin{itemize}
    \item EQ1: How do different PSI-PFL thresholds (Low, Medium-low, etc.) impact the global accuracy achieved across various client configurations?
    \item EQ2: Does PSI-PFL consistently achieve higher global accuracy than other baselines under varying levels of data heterogeneity?
    \item EQ3: How does PSI-PFL enhance client fairness compared to state-of-the-art baselines, particularly in highly heterogeneous data partitions?
\end{itemize}

Each of the previous EQs is answered using the correspondent observation highlighted in a gray box.

\subsection{Experimental Details Design}

\paragraph{\textbf{Testbed.}}
Our experiments were executed using a machine with 222 GB of disk, an AMD Ryzen Threadripper PRO 5995WX 64-cores CPU @ 3.44 GHz processors, a RAM of 500.0 GB, a Linux beastbianv2 6.1.0-20-amd64 OS, a GPU NVIDIA RTX A6000 50GB and Python 3.10. The FL models were trained using the Flower~\cite{beutel2020flower} and FedLab~\cite{zeng2023fedlab} frameworks.

\paragraph{\textbf{Data and partition protocol.}} We evaluate PSI-PFL across various datasets (tabular, image, and text), models, and data distributions. The details of the centralized datasets are summarized in Table~\ref{tab:characteristics_datasets}. These datasets were chosen due to their extensive use in prior research on non-IID data in FL. We employed a widely used partitioning protocol to simulate varying data distributions: Dirichlet~\cite{jimenez2024fedartml}. This protocol controls the degree of non-IIDness through a single parameter ($\alpha$), where smaller parameter values result in more heterogeneous client distributions. Additional details about the strategy used to divide the datasets among clients are provided in the corresponding subsection of Appendix~\ref{app:A}.


\begin{table}[htbp]
    \centering
    \caption{Characteristics of the datasets}
    \resizebox{\textwidth}{!}{
    \begin{tabular}{ccccc}
    \hline
    \textbf{Dataset} & \textbf{Modality} & \makecell{\textbf{Task}} & \textbf{\# classes} & \textbf{\# examples} \\
    \hline
    ACS Income~\cite{ding2021retiring} & Tabular & Classify high/low salary  & 2 & 1,664,500 \\
    Dutch~\cite{van20012001} & Tabular & Classify high/low salary  & 2 & 60,420 \\
    CelebA~\cite{liu2015deep} & Image & Detect smiling & 2 & 202,599 \\
    Sent140~\cite{go2009twitter} & Text & Sentiment analysis & 3 & 1,6000,000 \\
    \hline
    \end{tabular}}
    \label{tab:characteristics_datasets}
\end{table}

\paragraph{\textbf{Selection of $\alpha$ values.}}
The $\alpha$ values were chosen to cover the entire spectrum of non-IID settings, ranging from 0 (extreme non-IID) to 100 (IID). We tested eleven $\alpha$ values within this range, but, for brevity, present results for $\alpha = {0.3, 0.7, 50}$, as they are representative and consistent with the overall findings.


\paragraph{\textbf{Models.}} Tables~\ref{tab:acs_dutch},~\ref{tab:celeba}, and ~\ref{tab:sent140} in Appendix~\ref{app:B}, present detailed information about the model architecture employed in our experiments. For Dutch and ACS Income, we used a simple neural network with a single linear layer to replicate a Logistic Regression. 
For CelebA, we use a convolutional NN (CNN) with three convolutional layers (8, 16, and 32 channels) followed by max pooling and a fully connected layer with 2048 hidden units using ReLU activation.  For Sent140, we utilized a sequential NN with an embedding layer, a dropout layer (rate 0.5), an LSTM layer (10 units, dropout 0.2, recurrent dropout 0.2), and a dense output layer with softmax activation. We run the experiments using five different data partitions every time (i.e., five random seeds) to get more robust results.


\paragraph{\textbf{Metrics.}}\label{subsec:metrics}
This section describes the set of metrics that are often considered in our experiments. \\
\noindent
\textit{Accuracy.} It refers to the proportion of correctly classified instances compared to the total data size. Higher values of accuracy indicate a better model performance.
Because in FL, data is distributed among client devices, the accuracy can computed using either a global or a local perspective.

\noindent
\textit{Local accuracy}. Formally, local accuracy is computed individually for each client $k$ as follows:
$A_k=\frac{C_k}{n_k}$ 
where $C_k$ indicates the number of correctly classified samples on client $k$.
\\
\noindent
\textit{Global accuracy}. The global accuracy can be calculated as the weighted average of individual client accuracies, where the weight corresponds to the number of samples held by each client:
$A_{global} = \frac{\sum_{k=1}^{K} n_k A_k}{\sum_{k=1}^{K} n_k}$

where:
\begin{enumerate}
    \item $K$ is the total number of participating clients.
    \item $n_k$ is the number of data samples on client $k$.
\end{enumerate}

\noindent
\textit{Client fairness}. A popular definition of fairness in FL is \textit{Client Parity (CP)}, which requires that all FL clients achieve similar accuracy values (or loss values)~\cite{shi2023towards}.
CP is often quantified as the maximum relative performance difference between clients:
$CP=\frac{\max_{i,j=1}^K |{\phi_i} - {\phi_j}|}{\min_{i=1}^K {\phi_i}}$.
Where $\phi_i$ corresponds to the local accuracy of the \textit{$i^{th}$} client. In this case, smaller values of CP indicate higher client fairness.

We employ additional metrics to measure each solution's difference from the CL (target). For such purpose, we calculate the average distance $AD=\frac{1}{K} \sum_{i=1}^K (|\phi_i-\phi_{CL}|)$ and the correspondent standard deviation as $SDAD=\frac{1}{K} \sum_{i=1}^K (\phi_i - \phi_{CL})^2$. In both cases, $\phi_{CL}$ represents the accuracy of the CL (target) method. Smaller AD values indicate a smaller distance to the CL (target), and smaller SDAD indicates a smaller variation (desired) in such distances.

\subsection{PSI threshold impact}

\begin{figure}[ht]
\centering
\includegraphics[width=0.99\textwidth]{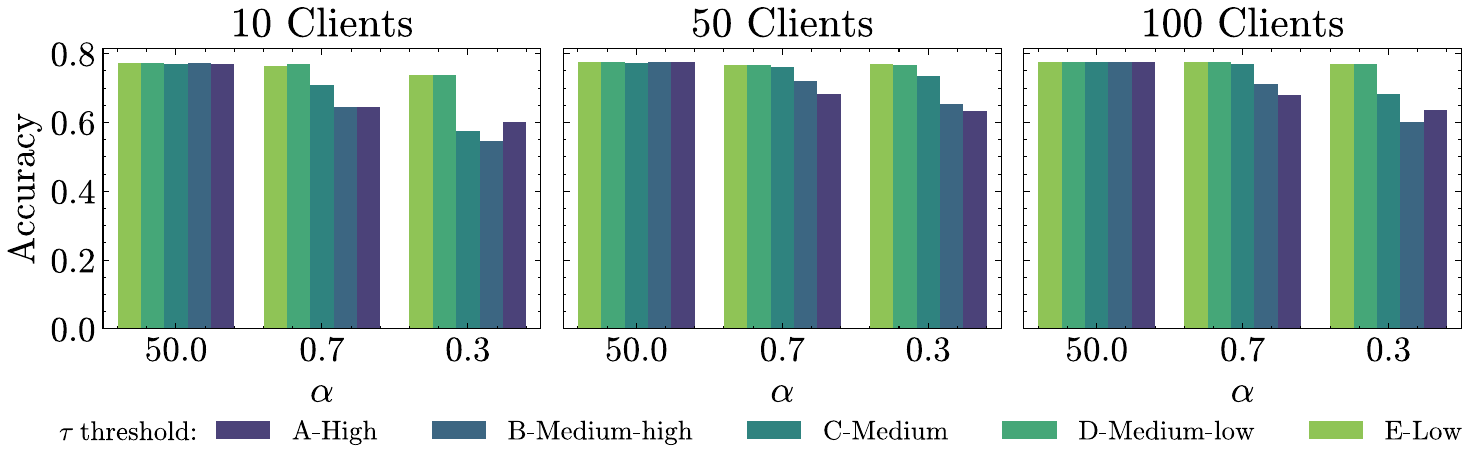}
\caption{Global test accuracy for PSI-PFL thresholds ($\tau$) for the ACSIncome dataset.}
\label{fig:acsincome_psi_thr_acc_dirichlet_all_CLI_LS_all_Alp}
\end{figure}

In this subsection, we answer EQ1 by examining the impact of the $\tau$ threshold on PSI-based client selection determined via the systematic approach described in Section~\ref{sec:strategy}. Fig.~\ref{fig:acsincome_psi_thr_acc_dirichlet_all_CLI_LS_all_Alp} illustrates the global accuracy of PSI-PFL as a function of the number of clients and the $\alpha$ parameter controlling non-IIDness. Notably, under medium to high levels of non-IIDness, the optimal $\tau$ values are typically in the lower to medium-low range (10th and 25th PSI percentiles). Furthermore, increasing the number of clients (e.g., 50 or 100) significantly enhances PSI-PFL's utility. The results in the following sections are based on the optimal $\tau$ threshold. The latter is consistent across all the datasets as depicted in Fig.~\ref{fig:dutch_psi_thr_acc_dirichlet_all_CLI_LS_all_Alp}, ~\ref{fig:celeba_psi_thr_acc_dirichlet_all_CLI_LS_all_Alp}, and ~\ref{fig:sent140_psi_thr_acc_dirichlet_all_CLI_LS_all_Alp} of  Appendix~\ref{app:C}.

\mybox{gray}{\observation{Using Low and Medium-low PSI-PFL thresholds facilitates obtaining higher global accuracy for all client and alpha configurations.}}

\subsection{PSI-PFL performance evaluation}

This subsection evaluates the PSI-PFL's global and local performance against the baselines.

\paragraph{\textbf{Global-level test PSI-PFL behavior.}}
At the global level, FL aggregates model updates from multiple distributed clients to produce a centralized model that generalizes across the entire dataset. 
This perspective evaluates overall model performance, including metrics such as accuracy, convergence rate, and robustness to non-IID data distributions.
The global results provide insights into the effectiveness of the FL framework as a whole, reflecting how well the aggregated model performs when deployed across all clients. 

\begin{figure}[ht]
\centering
\includegraphics[width=0.99\textwidth]{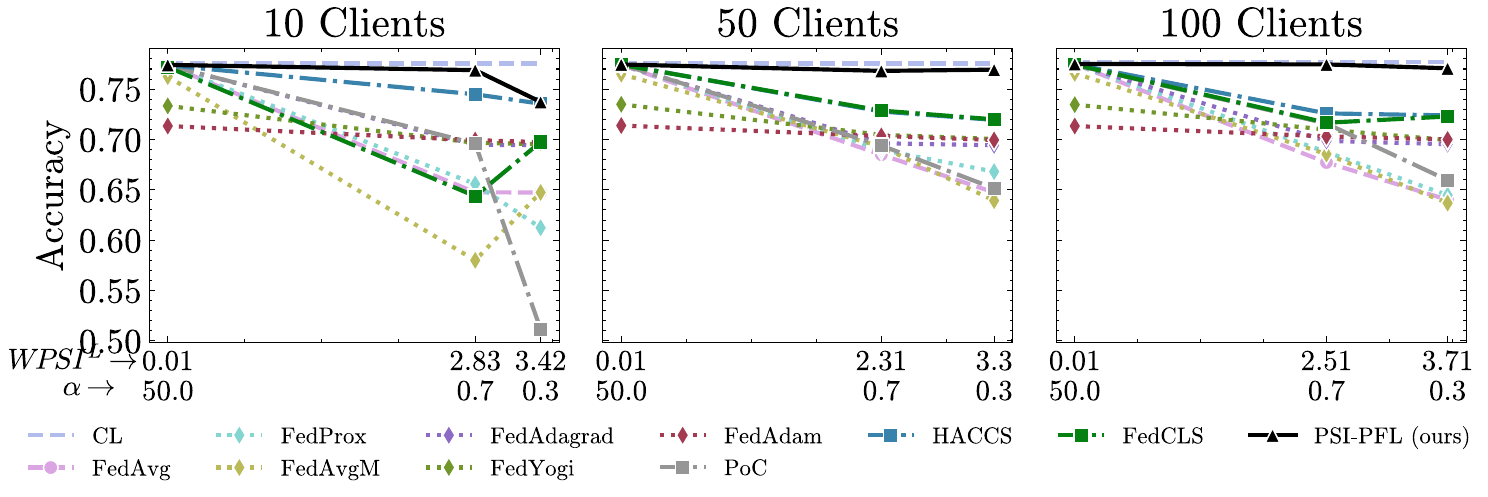}
\caption{Global test accuracy for all baselines and ACSIncome dataset.}
\label{fig:acsincome_global_acc_dirichlet_all_CLI_LS_all_Alp}
\end{figure}

We aim to answer EQ2 using Fig.~\ref{fig:acsincome_global_acc_dirichlet_all_CLI_LS_all_Alp}. It shows the global accuracy of PSI-PFL and all baselines varying the number of clients and the $\alpha$ parameter, along with the corresponding $WPSI^L$. PSI-PFL consistently outperforms the baselines, even under high non-IIDness, in up to 10\% relative increase in global accuracy compared to HACCS and FedCLS, the most challenging competitors. Its performance improves in scenarios with more clients (e.g., 50 or 100). The latter is consistent across all the datasets as depicted in Fig.~\ref{fig:dutch_global_acc_dirichlet_all_CLI_LS_all_Alp},~\ref{fig:celeba_global_acc_dirichlet_all_CLI_LS_all_Alp}, and ~\ref{fig:sent140_global_acc_dirichlet_all_CLI_LS_all_Alp} of Appendix~\ref{app:C}.

\mybox{gray}{\observation{
PSI-PFL systematically yields greater global accuracy than the other baselines for all levels of data heterogeneity (non-IIDness).
}}

\paragraph{\textbf{Local-level test PSI-PFL behavior.}}

%
Local-level results focus on the performance of individual client models before and after federated updates.
These results highlight variations due to heterogeneous data distributions, computational constraints, and personalization needs. %
By analyzing local outcomes, we can assess the impact of FL on individual nodes, ensuring fairness and identifying potential disparities in model efficacy across different clients.

\begin{figure}[ht]
\centering
\includegraphics[width=0.99\textwidth]{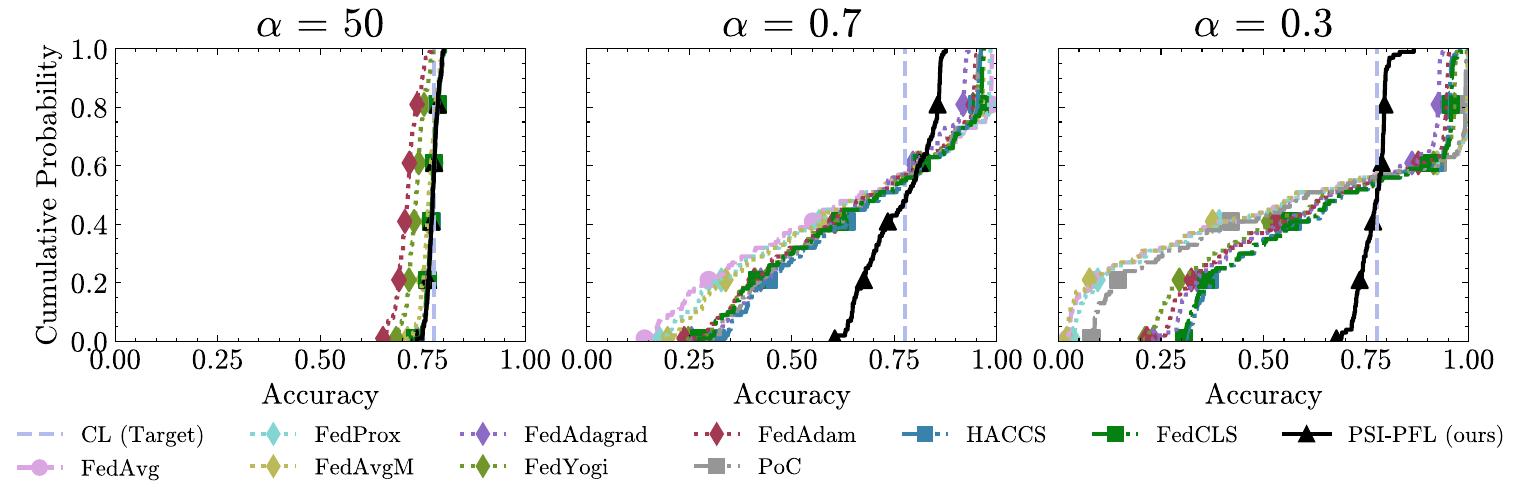}
\caption{Local test accuracy ECDF for all baselines, ACSIncome dataset, and 100 clients.}
\label{fig:acsincome_local_acc_dirichlet_100_CLI_LS_all_Alp}
\end{figure}

EQ3 is answered based on the results in Fig.~\ref{fig:acsincome_local_acc_dirichlet_100_CLI_LS_all_Alp}. It presents the empirical cumulative distribution function (ECDF) of local accuracy for PSI-PFL and all baselines across varying $\alpha$ values. Notably, the PSI-PFL distribution aligns more closely with the CL (target) distribution than the baselines, highlighting its superior performance in terms of client fairness. Our PS-PFL solution achieved a $CP=0.27$  under $\alpha=0.3$, the fairest value compared to the other baselines, which obtained CPs of the order of 2.2 for solutions such as HACCS and FedCLS and 3.7 for FedYogi, FedAdam, and FedAdagrad. Under the most heterogeneous scenario, our solution achieves an AD of 3\% from the CL (target), with an SDAD of 6\%. In comparison, baseline methods like HACCS and FedCLS have an AD of 24\% with an SDAD of 13\%. Such behavior corroborates a much closer distance of PSI-PFL to the CL (target) with less variability. The latter results are consistent across all the datasets as showcased in Fig.~\ref{fig:acsincome_three_images_vertical},~\ref{fig:dutch_three_images_vertical},~\ref{fig:celeba_three_images_vertical}, and~\ref{fig:sent140_three_images_vertical} of Appendix~\ref{app:C}.

\mybox{gray}{\observation{PSI-PFL fosters client fairness, with a more significantly small average distance of 3\% from the CL (target), outperforming the state-of-the-art baselines in high heterogeneity data partitions.}}


\section{Conclusion and Future Work}
\label{sec:conclusion}

In this work, we propose the PSI-PFL framework, which leverages the PSI as a novel metric for client selection in non-IID PFL. Our findings demonstrate that PSI effectively quantifies data heterogeneity in multiple data modalities (tabular, image, text), enabling the selection of more homogeneous clients and significantly improving global and local model performance even under highly non-IID scenarios. Compared to existing non-IID mitigation techniques, PSI-PFL consistently outperforms state-of-the-art baselines by up to 10\%  relative increase in global accuracy while ensuring fairer local performance, reaching only a 3\% average difference compared to the CL (target), making it a valuable tool for enhancing decentralized learning in heterogeneous environments.

In future research, we will explore PSI's applicability to other forms of data skew beyond label distribution, such as attribute or quantity skew. Employing more complex datasets with more label classes will help assess PSI-PFL's scalability and effectiveness in diverse settings. Although PSI-PFL surpasses state-of-the-art solutions in global-level performance on non-IID data, there is still room for improvement at the local level, ensuring a fairer distribution of performance across all clients and achieving closer alignment with the CL (target) behavior. Testing the results of PSI-PFL using alternative partitioning methods such as Similarity~\cite{wang2023distribution} is an interesting future avenue to prove its effectiveness under different data dispositions. Lastly, this study did not investigate the implementation of privacy techniques such as Differential Privacy (DP)~\cite{erlingsson2014rappor} or Multiparty Computing (MPC)~\cite{bohler2021secure} to enhance the privacy of the proposed FL solution, as it falls outside the scope of this work, remaining a promising avenue for future research.

\section{Acknowledgements}
\label{sec:ack}
To be included after acceptance.

\bibliographystyle{abbrv}
\bibliography{references}

\begin{thebibliography}{10}

\bibitem{ahmed2023semisupervised}
U.~Ahmed, J.~C.-W. Lin, and G.~Srivastava.
\newblock Semisupervised federated learning for temporal news hyperpatism detection.
\newblock {\em IEEE Transactions on Computational Social Systems}, 10(4):1758--1769, 2023.

\bibitem{beutel2020flower}
D.~J. Beutel, T.~Topal, A.~Mathur, X.~Qiu, J.~Fernandez-Marques, Y.~Gao, L.~Sani, K.~H. Li, T.~Parcollet, P.~P.~B. de~Gusm{\~a}o, et~al.
\newblock Flower: A friendly federated learning research framework.
\newblock {\em arXiv preprint arXiv:2007.14390}, 2020.

\bibitem{bohler2021secure}
J.~B{\"o}hler and F.~Kerschbaum.
\newblock Secure multi-party computation of differentially private heavy hitters.
\newblock In {\em Proceedings of the 2021 ACM SIGSAC Conference on Computer and Communications Security}, pages 2361--2377, 2021.

\bibitem{chen2022emd}
A.~Chen, Y.~Fu, Z.~Sha, and G.~Lu.
\newblock An emd-based adaptive client selection algorithm for federated learning in heterogeneous data scenarios.
\newblock {\em Frontiers in Plant Science}, 13:908814, 2022.

\bibitem{apple_privacy_2017}
{Differential Privacy Team}.
\newblock Learning with privacy at scale.
\newblock \url{https://machinelearning.apple.com/research/learning-with-privacy-at-scale}, 2017.
\newblock Accessed: 2024-10-16.

\bibitem{ding2021retiring}
F.~Ding, M.~Hardt, J.~Miller, and L.~Schmidt.
\newblock Retiring adult: New datasets for fair machine learning.
\newblock {\em Advances in neural information processing systems}, 34:6478--6490, 2021.

\bibitem{du2023proposed}
J.~du~Pisanie, J.~S. Allison, and J.~Visagie.
\newblock A proposed simulation technique for population stability testing in credit risk scorecards.
\newblock {\em Mathematics}, 11(2):492, 2023.

\bibitem{erlingsson2014rappor}
{\'U}.~Erlingsson, V.~Pihur, and A.~Korolova.
\newblock Rappor: Randomized aggregatable privacy-preserving ordinal response.
\newblock In {\em Proceedings of the 2014 ACM SIGSAC conference on computer and communications security}, pages 1054--1067, 2014.

\bibitem{g2024noniiddatafederatedlearning}
D.~M.~J. G., D.~Solans, M.~Heikkila, A.~Vitaletti, N.~Kourtellis, A.~Anagnostopoulos, and I.~Chatzigiannakis.
\newblock Non-iid data in federated learning: A survey with taxonomy, metrics, methods, frameworks and future directions, 2024.

\bibitem{go2009twitter}
A.~Go, R.~Bhayani, and L.~Huang.
\newblock Twitter sentiment classification using distant supervision.
\newblock {\em CS224N project report, Stanford}, 1(12):2009, 2009.

\bibitem{goussakov2020hellinger}
R.~Goussakov.
\newblock Hellinger distance-based similarity measures for recommender systems, 2020.

\bibitem{hsieh2020non}
K.~Hsieh, A.~Phanishayee, O.~Mutlu, and P.~Gibbons.
\newblock The non-iid data quagmire of decentralized machine learning.
\newblock In {\em International Conference on Machine Learning}, pages 4387--4398. PMLR, 2020.

\bibitem{hsu2019measuring}
T.-M.~H. Hsu, H.~Qi, and M.~Brown.
\newblock Measuring the effects of non-identical data distribution for federated visual classification.
\newblock {\em arXiv preprint arXiv:1909.06335}, 2019.

\bibitem{pmlr-v151-jee-cho22a}
Y.~Jee~Cho, J.~Wang, and G.~Joshi.
\newblock Towards understanding biased client selection in federated learning.
\newblock In G.~Camps-Valls, F.~J.~R. Ruiz, and I.~Valera, editors, {\em Proceedings of The 25th International Conference on Artificial Intelligence and Statistics}, volume 151 of {\em Proceedings of Machine Learning Research}, pages 10351--10375. PMLR, 2022.

\bibitem{jimenez2024fedartml}
G.~D.~M. Jimenez, A.~Anagnostopoulos, I.~Chatzigiannakis, and A.~Vitaletti.
\newblock Fedartml: A tool to facilitate the generation of non-iid datasets in a controlled way to support federated learning research.
\newblock {\em IEEE Access}, 2024.

\bibitem{li2022fedcls}
C.~Li and H.~Wu.
\newblock {FedCLS}: A federated learning client selection algorithm based on cluster label information.
\newblock In {\em 2022 IEEE 96th Vehicular Technology Conference (VTC2022-Fall)}, pages 1--5. IEEE, 2022.

\bibitem{aggregators_FL}
Q.~Li, Y.~Diao, Q.~Chen, and B.~He.
\newblock Federated learning on non-iid data silos: An experimental study.
\newblock In {\em 2022 IEEE 38th International Conference on Data Engineering (ICDE)}, pages 965--978, IEEE, 2022. IEEE, IEEE.

\bibitem{li2020federated}
T.~Li, A.~K. Sahu, A.~Talwalkar, and V.~Smith.
\newblock Federated learning: Challenges, methods, and future directions.
\newblock {\em IEEE signal processing magazine}, 37(3):50--60, 2020.

\bibitem{lin1991divergence}
J.~Lin.
\newblock Divergence measures based on the shannon entropy.
\newblock {\em IEEE Transactions on Information theory}, 37(1):145--151, 1991.

\bibitem{liu2015deep}
Z.~Liu, P.~Luo, X.~Wang, and X.~Tang.
\newblock Deep learning face attributes in the wild.
\newblock In {\em Proceedings of the IEEE international conference on computer vision}, pages 3730--3738, 2015.

\bibitem{mcmahan2017communication}
B.~McMahan, E.~Moore, D.~Ramage, S.~Hampson, and B.~A. y~Arcas.
\newblock Communication-efficient learning of deep networks from decentralized data.
\newblock In {\em Artificial intelligence and statistics}, pages 1273--1282. PMLR, 2017.

\bibitem{pei2024review}
J.~Pei, W.~Liu, J.~Li, L.~Wang, and C.~Liu.
\newblock A review of federated learning methods in heterogeneous scenarios.
\newblock {\em IEEE Transactions on Consumer Electronics}, 2024.

\bibitem{reddi2020adaptive}
S.~Reddi, Z.~Charles, M.~Zaheer, Z.~Garrett, K.~Rush, J.~Kone{\v{c}}n{\`y}, S.~Kumar, and H.~B. McMahan.
\newblock Adaptive federated optimization.
\newblock {\em arXiv preprint arXiv:2003.00295}, 2020.

\bibitem{rubner2000earth}
Y.~Rubner, C.~Tomasi, and L.~J. Guibas.
\newblock The earth mover's distance as a metric for image retrieval.
\newblock {\em International journal of computer vision}, 40:99--121, 2000.

\bibitem{shi2023towards}
Y.~Shi, H.~Yu, and C.~Leung.
\newblock Towards fairness-aware federated learning.
\newblock {\em IEEE Transactions on Neural Networks and Learning Systems}, 2023.

\bibitem{siddiqi2012credit}
N.~Siddiqi.
\newblock {\em Credit risk scorecards: developing and implementing intelligent credit scoring}, volume~3.
\newblock John Wiley \& Sons, 2012.

\bibitem{tan2022towards}
A.~Z. Tan, H.~Yu, L.~Cui, and Q.~Yang.
\newblock Towards personalized federated learning.
\newblock {\em IEEE transactions on neural networks and learning systems}, 34(12):9587--9603, 2022.

\bibitem{tan2023privacy}
Q.~Tan, S.~Wu, and Y.~Tao.
\newblock Privacy-enhanced federated learning for non-iid data.
\newblock {\em Mathematics}, 11(19):4123, 2023.

\bibitem{van20012001}
P.~Van~der Laan.
\newblock The 2001 census in the netherlands: Integration of registers and surveys.
\newblock In {\em CONFERENCE AT THE CATHIE MARSH CENTRE.}, pages 1--24, 2001.

\bibitem{wang2023distribution}
Y.~Wang, Y.~Tong, Z.~Zhou, R.~Zhang, S.~J. Pan, L.~Fan, and Q.~Yang.
\newblock Distribution-regularized federated learning on non-iid data.
\newblock In {\em 2023 IEEE 39th International Conference on Data Engineering (ICDE)}, pages 2113--2125. IEEE, 2023.

\bibitem{wolfrath2022haccs}
J.~Wolfrath, N.~Sreekumar, D.~Kumar, Y.~Wang, and A.~Chandra.
\newblock {HACCS}: Heterogeneity-aware clustered client selection for accelerated federated learning.
\newblock In {\em 2022 IEEE International Parallel and Distributed Processing Symposium (IPDPS)}, pages 985--995. IEEE, 2022.

\bibitem{xu2024fblg}
Y.~Xu, Y.~Li, H.~Luo, X.~Fan, and X.~Liu.
\newblock Fblg: A local graph based approach for handling dual skewed non-iid data in federated learning.
\newblock In {\em Proceedings of the Thirty-Third International Joint Conference on Artificial Intelligence, IJCAI-24, K. Larson, Ed. International Joint Conferences on Artificial Intelligence Organization}, volume~8, pages 5289--5297, 2024.

\bibitem{yurdakul2018statistical}
B.~Yurdakul.
\newblock {\em Statistical properties of population stability index}.
\newblock Western Michigan University, 2018.

\bibitem{zeng2023fedlab}
D.~Zeng, S.~Liang, X.~Hu, H.~Wang, and Z.~Xu.
\newblock Fedlab: A flexible federated learning framework.
\newblock {\em Journal of Machine Learning Research}, 24(100):1--7, 2023.

\bibitem{zhao2023non}
H.~Zhao.
\newblock Non-iid quantum federated learning with one-shot communication complexity.
\newblock {\em Quantum Machine Intelligence}, 5(1):3, 2023.

\end{thebibliography}

\label{sec:appendix}
\newpage
\section{Appendix}
This section presents the appendix related to the data distribution analysis, the architecture of the models employed, and the results of global and local testing obtained with the alternative datasets used.

\appendix 
\renewcommand{\thesection}{\Alph{section}} 

\section{Data distribution} \label{app:A}
The experiments in the paper involved four datasets: ACSIncome~\cite{ding2021retiring}, Dutch~\cite{van20012001}, CelebA~\cite{liu2015deep}, and Sent140~\cite{go2009twitter}. These datasets were partitioned into $K$ clients using a Dirichlet-based protocol~\cite{jimenez2024fedartml}, where the $\alpha$ parameter was adjusted to artificially simulate non-IID data. This method allowed us to simulate a FL environment with varying levels of non-IIDness. In this section, we present the ECDF of the PSI computed for each dataset across different $\alpha$ values, highlighting the diverse distributions achieved through the partitioning protocol.


\begin{figure}[ht]
\centering
\includegraphics[width=1\textwidth]{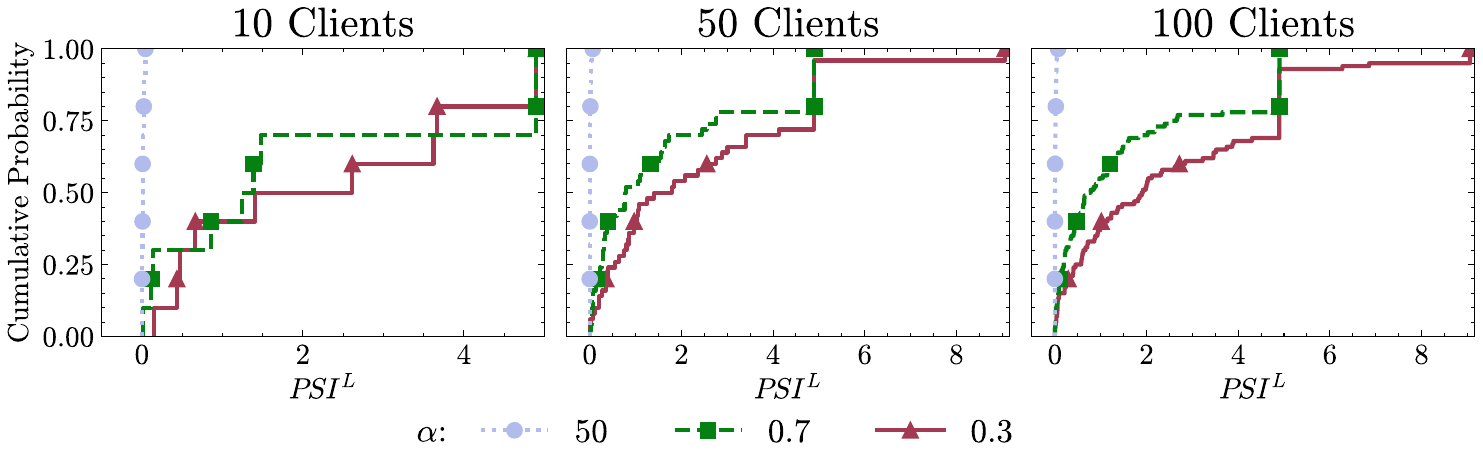}
\caption{Client's ECDF for the $PSI^L$ and the ACSIncome dataset.}
\label{fig:acsincome_psi_distr_dirichlet_all_CLI_LS_all_Alp}
\end{figure}



\begin{figure}[ht]
\centering
\includegraphics[width=1\textwidth]{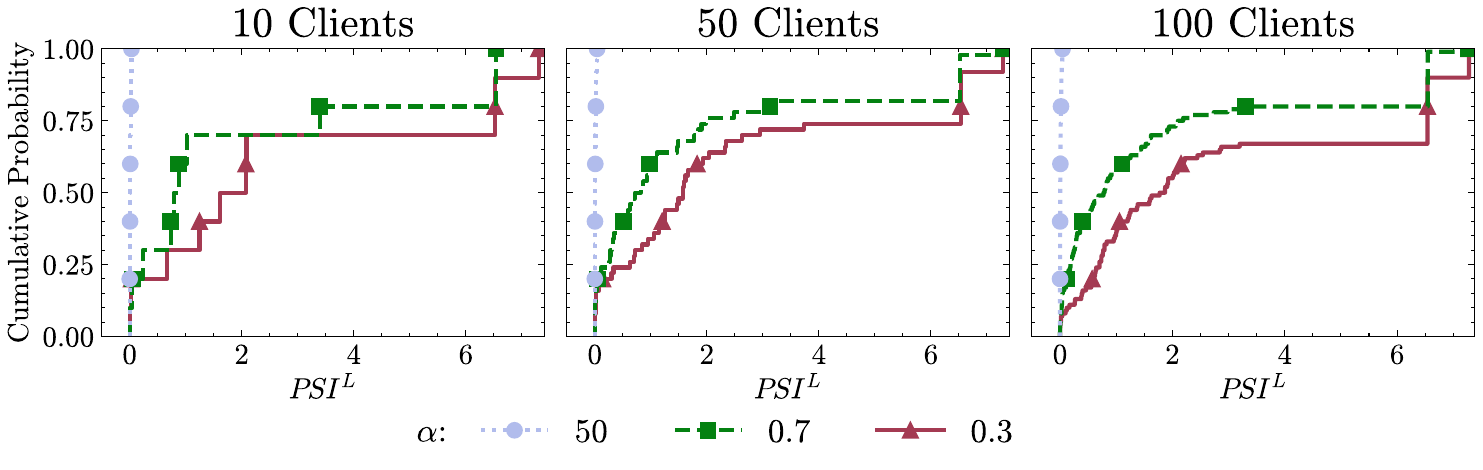}
\caption{Client's ECDF for the $PSI^L$ and the Dutch dataset.}
\label{fig:dutch_psi_distr_dirichlet_all_CLI_LS_all_Alp}
\end{figure}


\begin{figure}[ht]
\centering
\includegraphics[width=1\textwidth]{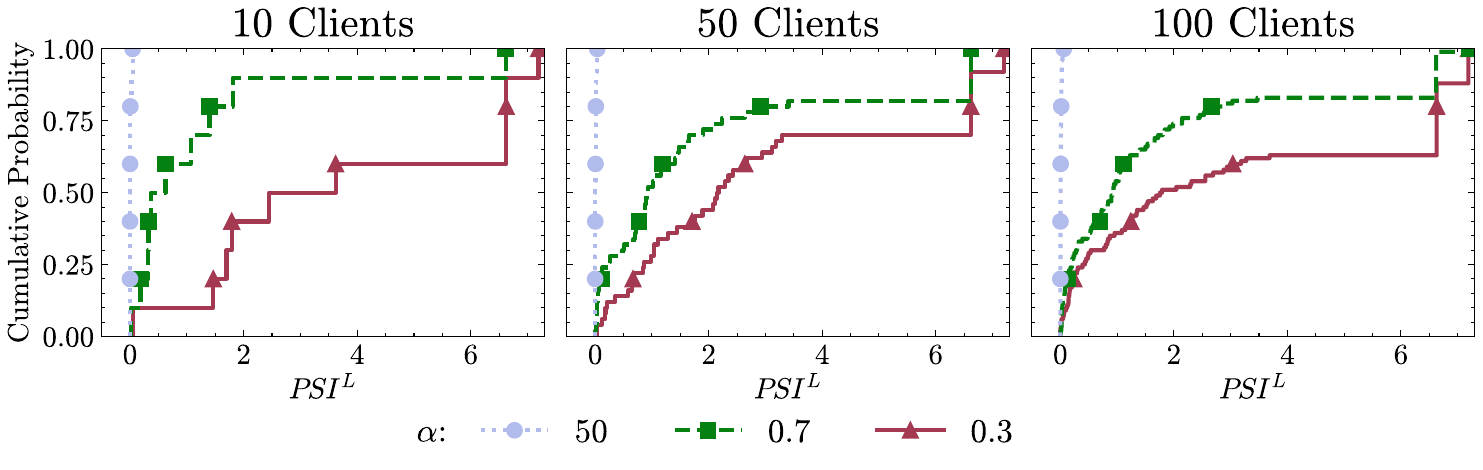}
\caption{Client's ECDF for the $PSI^L$ and the CelebA dataset.}
\label{fig:celeba_psi_distr_dirichlet_all_CLI_LS_all_Alp}
\end{figure}



\begin{figure}[ht]
\centering
\includegraphics[width=1\textwidth]{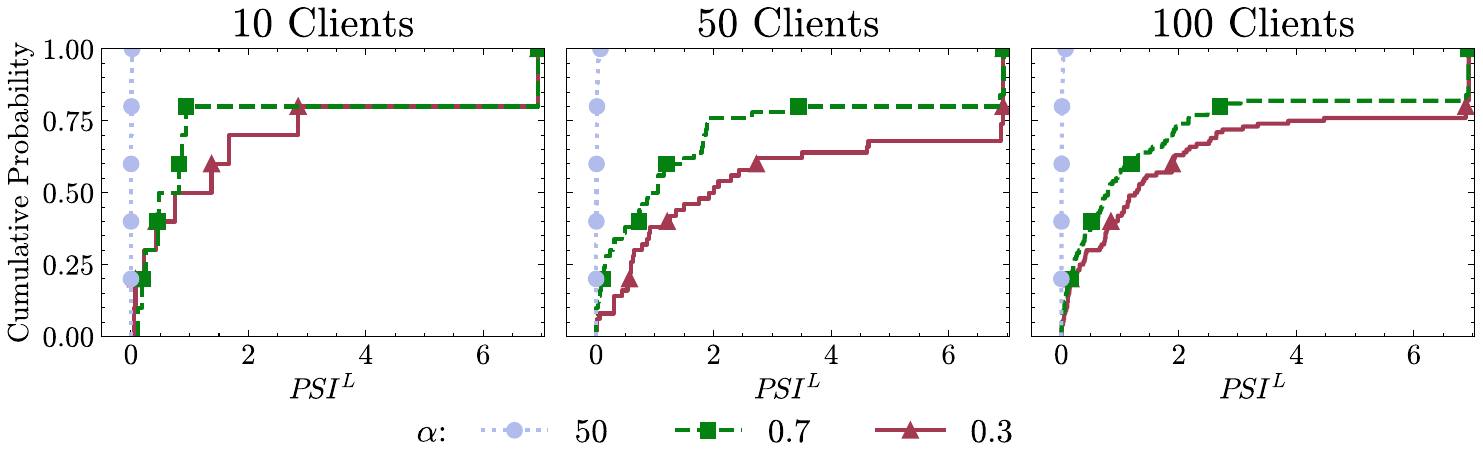}
\caption{Client's ECDF for the $PSI^L$ and the Sent140 dataset.}
\label{fig:sent140_psi_distr_dirichlet_all_CLI_LS_all_Alp}
\end{figure}

Figs.~\ref{fig:acsincome_psi_distr_dirichlet_all_CLI_LS_all_Alp}, ~\ref{fig:dutch_psi_distr_dirichlet_all_CLI_LS_all_Alp},~\ref{fig:celeba_psi_distr_dirichlet_all_CLI_LS_all_Alp}, and~\ref{fig:sent140_psi_distr_dirichlet_all_CLI_LS_all_Alp} illustrate the disruption of PSI values across clients using different Dirichlet partitioning settings for the four datasets, that corresponds to different heterogeneity of the partitioned clients. In all cases, a higher value of the parameter $\alpha$, which corresponds to a more IID partitioning, results in a steep ECDF curve, indicating a high concentration of PSI values that are very small (close to zero). On the other hand, a lower value of $\alpha$, which corresponds to a more non-IID partitioning, results in more diverse values of PSI, in which some clients exhibit significantly higher PSI values, indicating that data heterogeneity across the partitioned clients is reflected in the resulting PSI values. The figures demonstrate that the PSI value exhibits similar trends across the four datasets. They also show that for the high-heterogeneity settings ($\alpha = 0.7$ and $\alpha = 0.3$), increasing the level of non-IIDness leads to a distribution in which a more significant number of clients have higher PSI values. Finally, the figures highlight that the PSI can effectively quantify the level of non-IIDness in scenarios with varying numbers of clients (ranging from $10$ to $100$).

\section{Models architecture} \label{app:B}
The architectures of the models used in our experiments for each dataset are presented in Tables~\ref{tab:acs_dutch},~\ref{tab:celeba}, and~\ref{tab:sent140}.
 
\begin{table}[ht]
\centering
\begin{minipage}[t]{0.3\textwidth} 
    \centering
    \resizebox{\textwidth}{!}{%
        \begin{tabular}{cc}
        \hline
        \textbf{Layer} & \textbf{Description} \\ \hline 
        \makecell{Fully Connected \\ Layer} & (input size, 2) \\ \hline \\\\\\\\\\
        \end{tabular}
    }
    \caption{ACSIncome and Dutch} \label{tab:acs_dutch}
\end{minipage}\hfill
\begin{minipage}[t]{0.3\textwidth} 
    \centering
    \resizebox{\textwidth}{!}{%
        \begin{tabular}{cc}
        \hline
        \textbf{Layer} & \textbf{Description} \\ \hline
        Conv2D with Relu & (3, 8, 3, 1) \\ 
        Max Pooling & (2, 2) \\ 
        Conv2D with Relu & (8, 16, 3, 1) \\ 
        Max Pooling & (2, 2) \\ 
        Conv2D with Relu & (16, 32, 3, 1) \\ 
        Max Pooling & (2, 2) \\ 
        \makecell{Fully Connected \\ with Relu} & (8 $\times$ 8 $\times$ 32) \\ \hline
        \end{tabular}
    }
    \caption{CelebA} \label{tab:celeba}
\end{minipage}\hfill
\begin{minipage}[t]{0.3\textwidth} 
    \centering
    \resizebox{\textwidth}{!}{%
        \begin{tabular}{cc}
        \hline
        \textbf{Layer} & \textbf{Description} \\ \hline
        \makecell{GloVe 6B \\ Embeddings} & (25,300) \\ 
        LSTM & (25, 10) \\ 
        \makecell{Fully Connected \\ with Relu} & (10) \\ 
        \makecell{Fully Connected \\ with Softmax} & (2) \\ 
        \hline
        \end{tabular}
    }
    \caption{Sent140} \label{tab:sent140}
\end{minipage}

\end{table}

\section{Results with alternative datasets} \label{app:C}
Notice that the results obtained with the ACSIncome dataset in Section~\ref{sec:experiments} are consistent with the Dutch, CelebA, and Sent140 datasets. Therefore, in this appendix, we thoroughly analyze and discuss the results obtained in the Dutch dataset, and such results can be extrapolated to the other datasets by looking at the respective figures and subsections in this appendix.


\begin{figure}[htp]
    \centering
    \begin{subfigure}{\textwidth}
        \centering
        \includegraphics[width=1\textwidth]{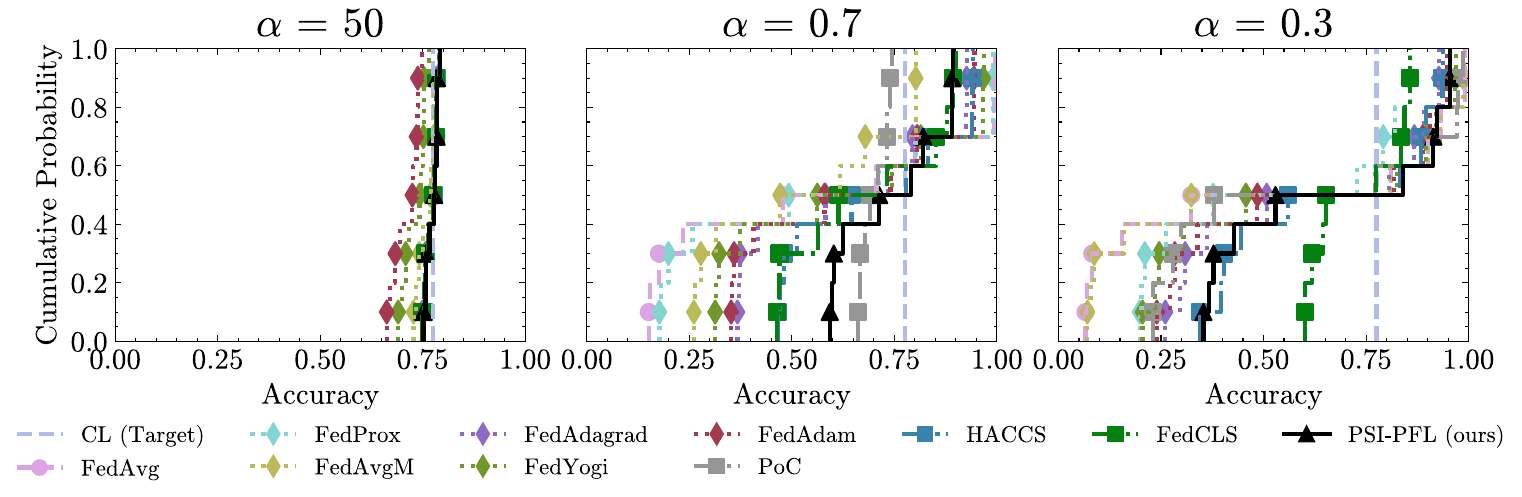}
        \caption{10 Clients}
        \label{fig:acsincome_local_acc_dirichlet_10_CLI_LS_all_Alp}
    \end{subfigure}
    
    \vspace{0.5cm} 
    
    \begin{subfigure}{\textwidth}
        \centering
        \includegraphics[width=1\textwidth]{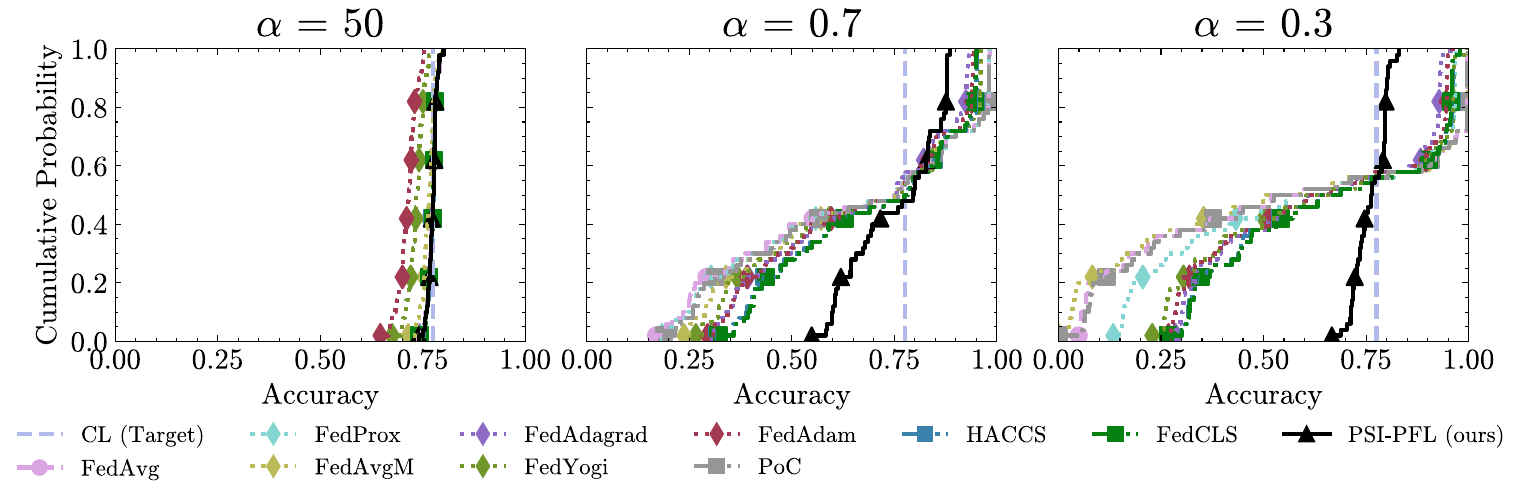}
        \caption{50 Clients}
        \label{fig:acsincome_local_acc_dirichlet_50_CLI_LS_all_Alp}
    \end{subfigure}
    
    \vspace{0.5cm} 
    
    \caption{Local test accuracy ECDF for all baselines for the ACSIncome dataset.}
    \label{fig:acsincome_three_images_vertical}
\end{figure}



\begin{figure}[ht]
\centering
\includegraphics[width=1\textwidth]{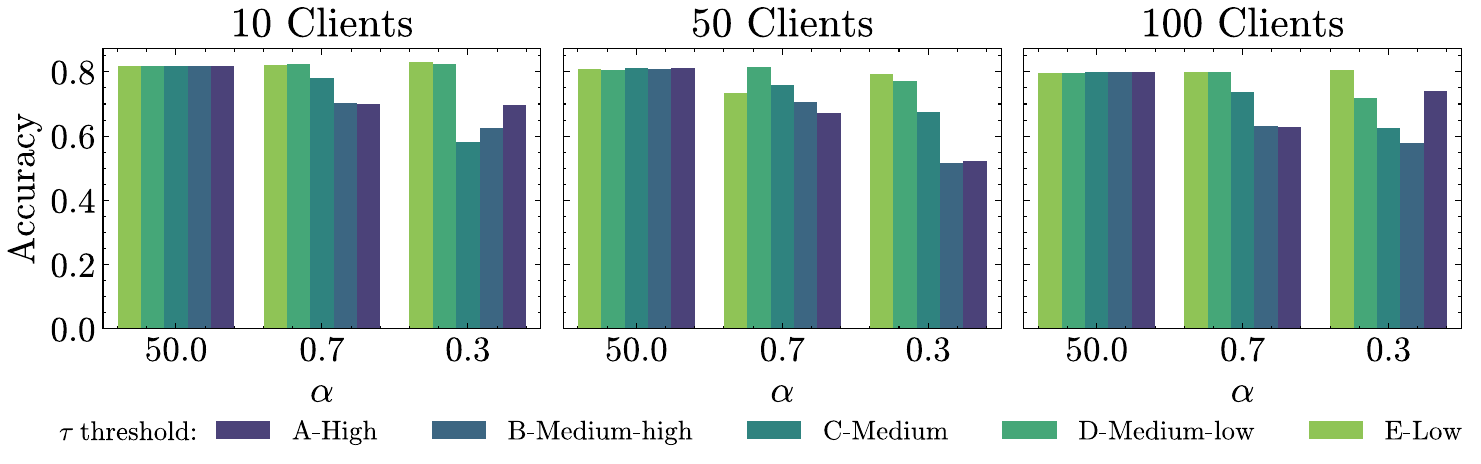}
\caption{Global test accuracy for PSI-PFL thresholds ($\tau$) for the Dutch dataset.}
\label{fig:dutch_psi_thr_acc_dirichlet_all_CLI_LS_all_Alp}
\end{figure}

As shown in Fig.~\ref{fig:dutch_psi_thr_acc_dirichlet_all_CLI_LS_all_Alp}, the overall behavior observed in the ACSIncome dataset is consistently maintained in the Dutch dataset.
As already observed, using high and medium-high $\tau$ values systematically yields better test accuracy values.
At the same time, we observe in high-heterogeneity settings (i.e., $\alpha=0.3$) with 100 clients, using low $\tau$ values yields better test accuracy than the medium and medium-low options.
This suggests that including all the clients during training (including those more heterogeneous) might have a positive effect as driven by the representativity of much more diverse clients.

\begin{figure}[ht]
\centering
\includegraphics[width=1\textwidth]{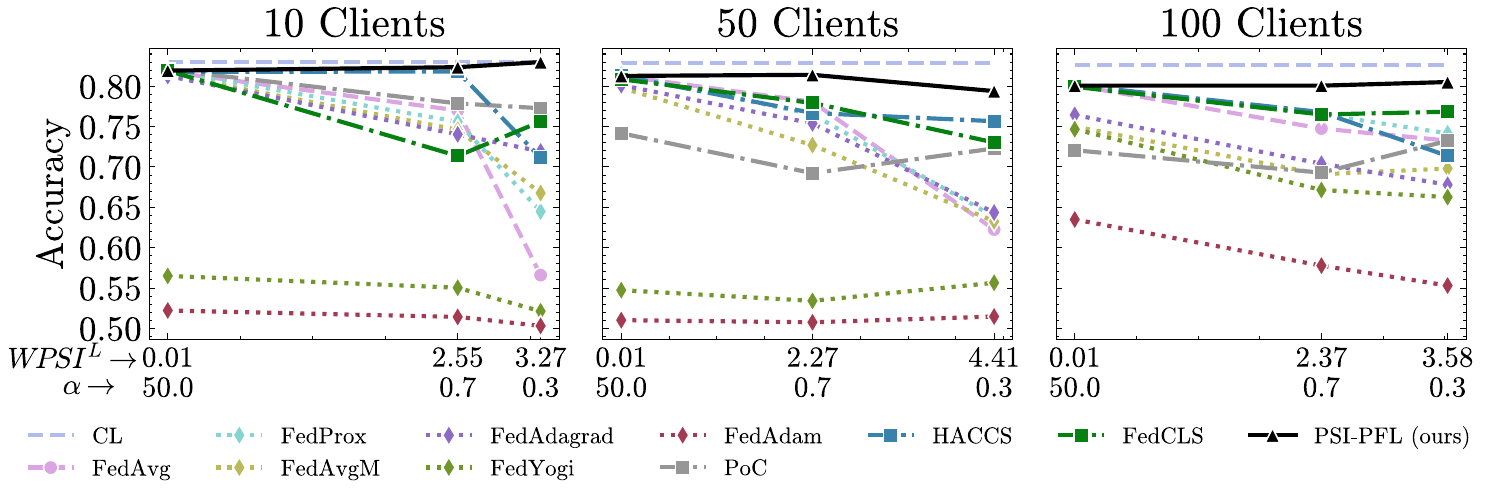}
\caption{Global test accuracy for all baselines and Dutch dataset.}
\label{fig:dutch_global_acc_dirichlet_all_CLI_LS_all_Alp}
\end{figure}

Fig.~\ref{fig:dutch_global_acc_dirichlet_all_CLI_LS_all_Alp} reports the global accuracy values obtained for all the considered baselines using the Dutch dataset.
As depicted there, PSI-FL is over the best-performing baselines for this dataset, with quite robust performance across heterogeneity values (driven by the $\alpha$ parameter).
Additionally, we see how FedAdam systematically performs worse than the rest of the baselines, with FedYogi yielding a performance that is very close to FedAdam but becomes much better (and closer to the rest of the baselines) for the 100 clients scenario.
While, in general, the performance of the baselines degrades with higher heterogeneity partitions, the performance degradation for higher $\alpha$ values exhibits an unpredictable pattern for most of the baselines.
In general, regularization-based solutions perform worse with higher heterogeneity levels.
Finally, we observe that PSI-PFL and baselines tend to have higher accuracy with more clients (i.e., 50, 100) than with fewer clients in the presence of high non-IDness levels.

\begin{figure}[htp]
    \centering
    \begin{subfigure}{\textwidth}
        \centering
        \includegraphics[width=1\textwidth]{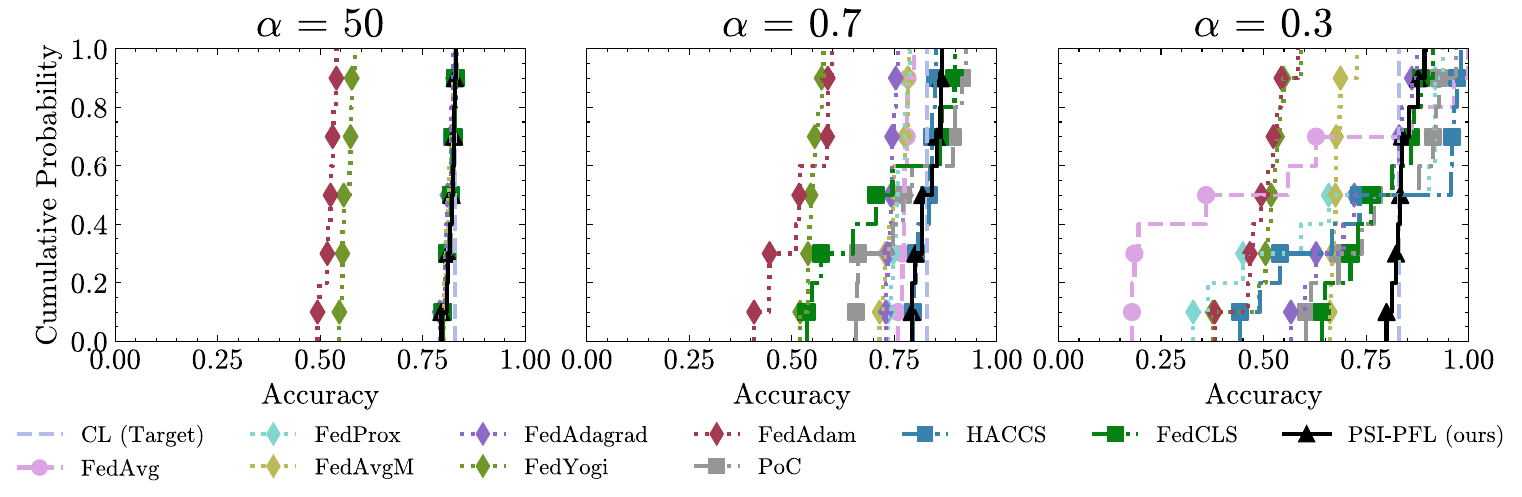}
        \caption{10 Clients}
        \label{fig:dutch_local_acc_dirichlet_10_CLI_LS_all_Alp}
    \end{subfigure}
    
    \vspace{0.5cm} 
    
    \begin{subfigure}{\textwidth}
        \centering
        \includegraphics[width=1\textwidth]{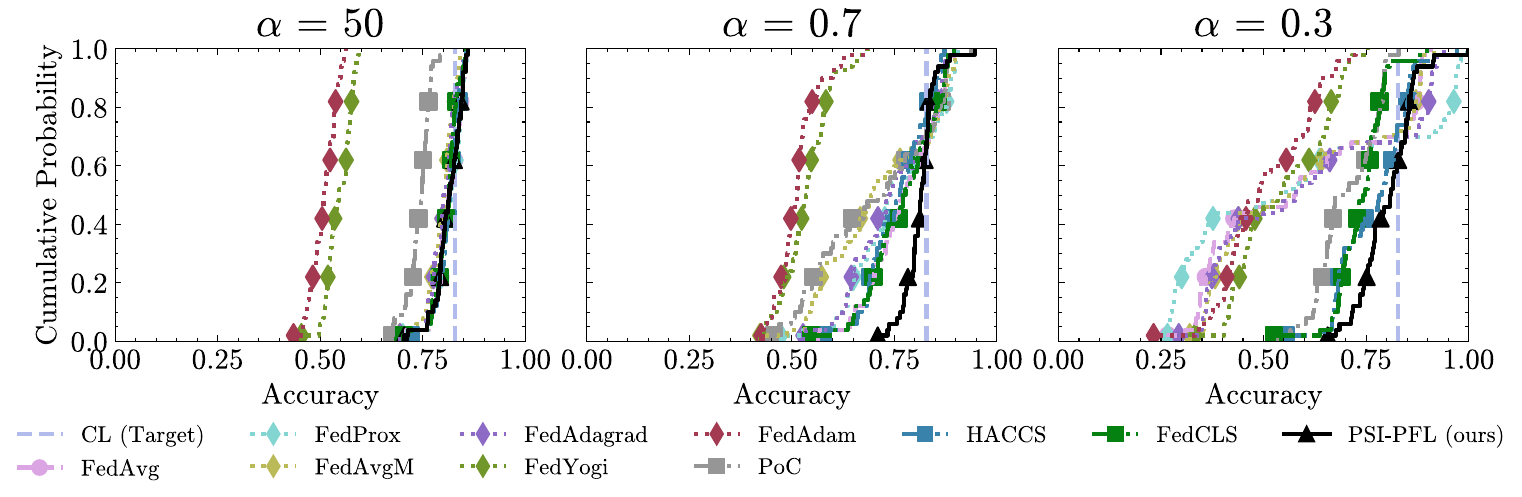}
        \caption{50 Clients}
        \label{fig:dutch_local_acc_dirichlet_50_CLI_LS_all_Alp}
    \end{subfigure}
    
    \vspace{0.5cm} 
    
    \begin{subfigure}{\textwidth}
        \centering
        \includegraphics[width=1\textwidth]{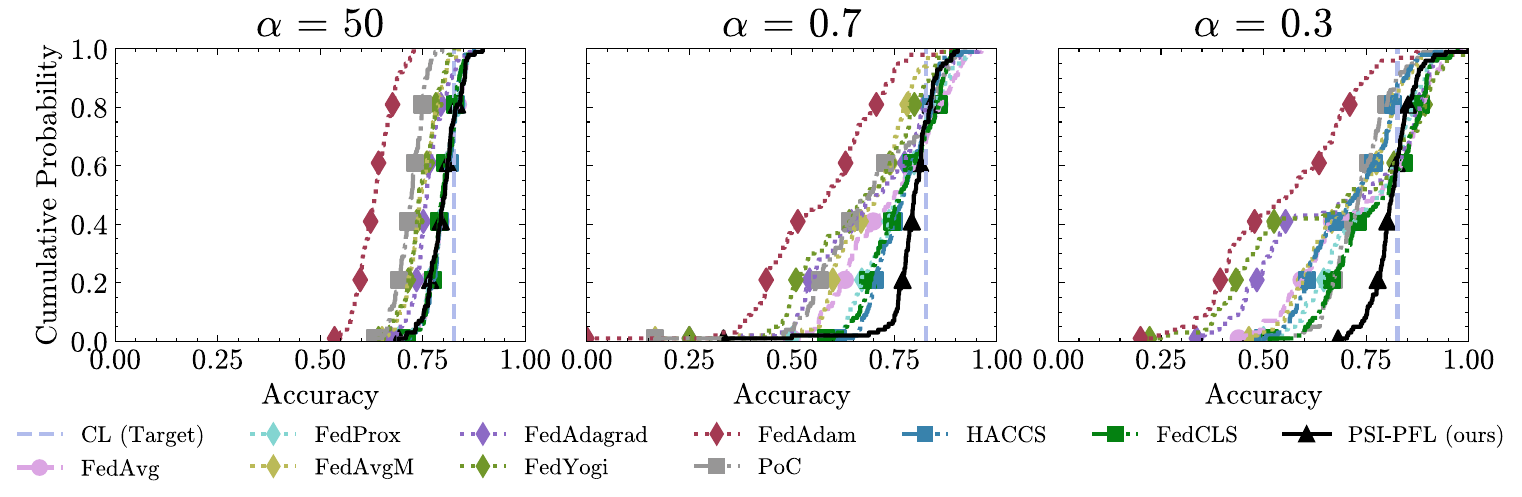}
        \caption{100 Clients}
        \label{fig:dutch_local_acc_dirichlet_100_CLI_LS_all_Alp}
    \end{subfigure}

    \caption{Local test accuracy ECDF for all baselines for the Dutch dataset.}
    \label{fig:dutch_three_images_vertical}
\end{figure}

PSI-FL demonstrates the highest average local performance among the techniques compared, highlighting its effectiveness in federated learning scenarios. 
Additionally, PSI-FL outperforms the best-performing baselines on this dataset, particularly in client fairness, except for 100 clients with $\alpha=0.7$.
Furthermore, under high data heterogeneity conditions ($\alpha=0.3$), PSI-PFL achieves high accuracy and effectively reduces local performance disparities among clients, ensuring a more balanced model performance across the federation.


\begin{figure}[ht]
\centering
\includegraphics[width=1\textwidth]{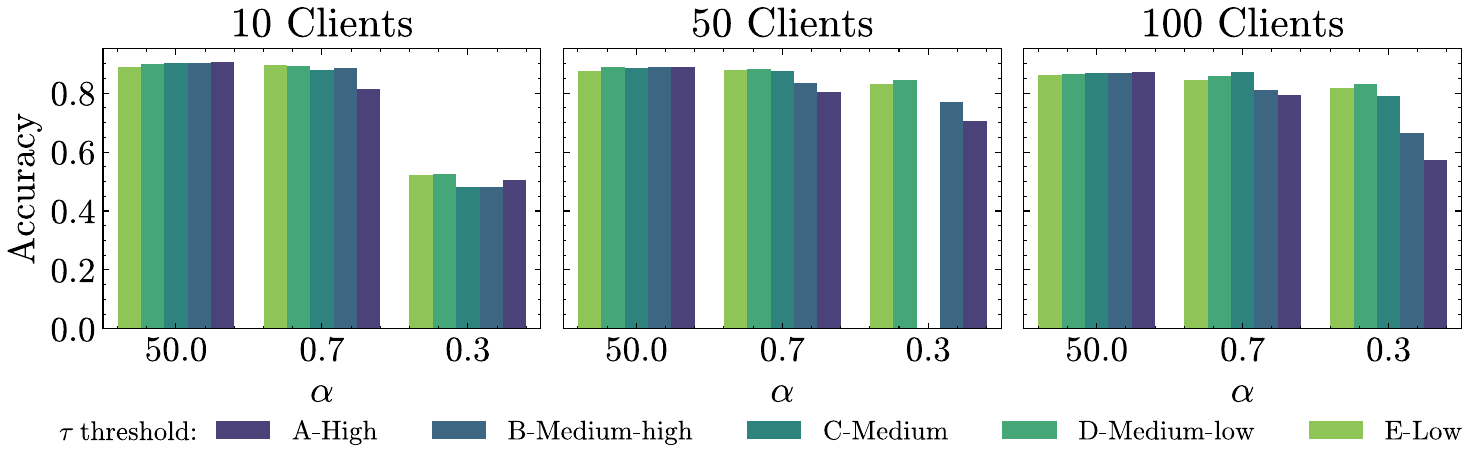}
\caption{Global test accuracy for PSI-PFL thresholds ($\tau$) for the CelebA dataset.}
\label{fig:celeba_psi_thr_acc_dirichlet_all_CLI_LS_all_Alp}
\end{figure}

\begin{figure}[ht]
\centering
\includegraphics[width=1\textwidth]{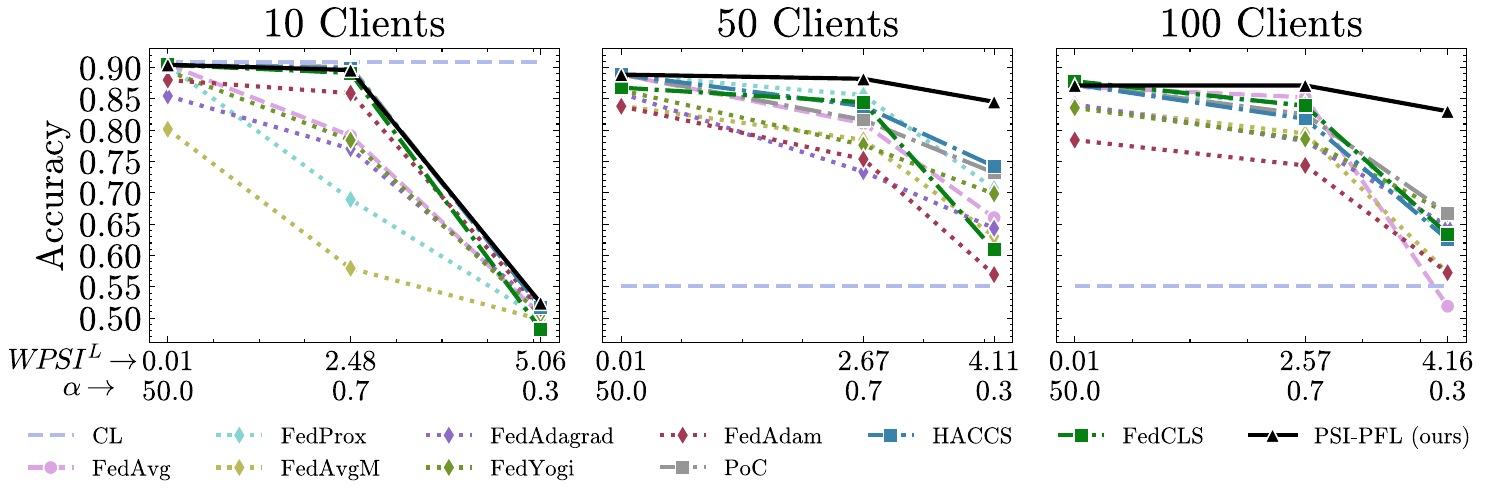}
\caption{Global test accuracy for all baselines and CelebA dataset.}
\label{fig:celeba_global_acc_dirichlet_all_CLI_LS_all_Alp}
\end{figure}

\begin{figure}[htp]
    \centering
    \begin{subfigure}{\textwidth}
        \centering
        \includegraphics[width=1\textwidth]{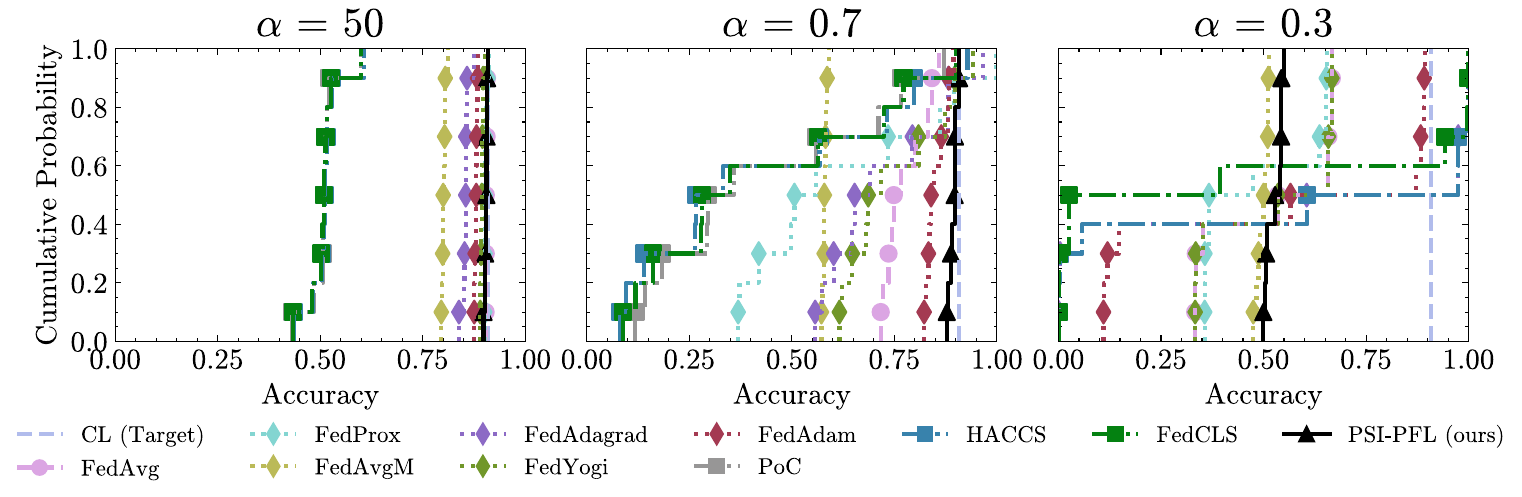}
        \caption{10 Clients}
        \label{fig:celeba_local_acc_dirichlet_10_CLI_LS_all_Alp}
    \end{subfigure}
    
    \vspace{0.5cm} 
    
    \begin{subfigure}{\textwidth}
        \centering
        \includegraphics[width=1\textwidth]{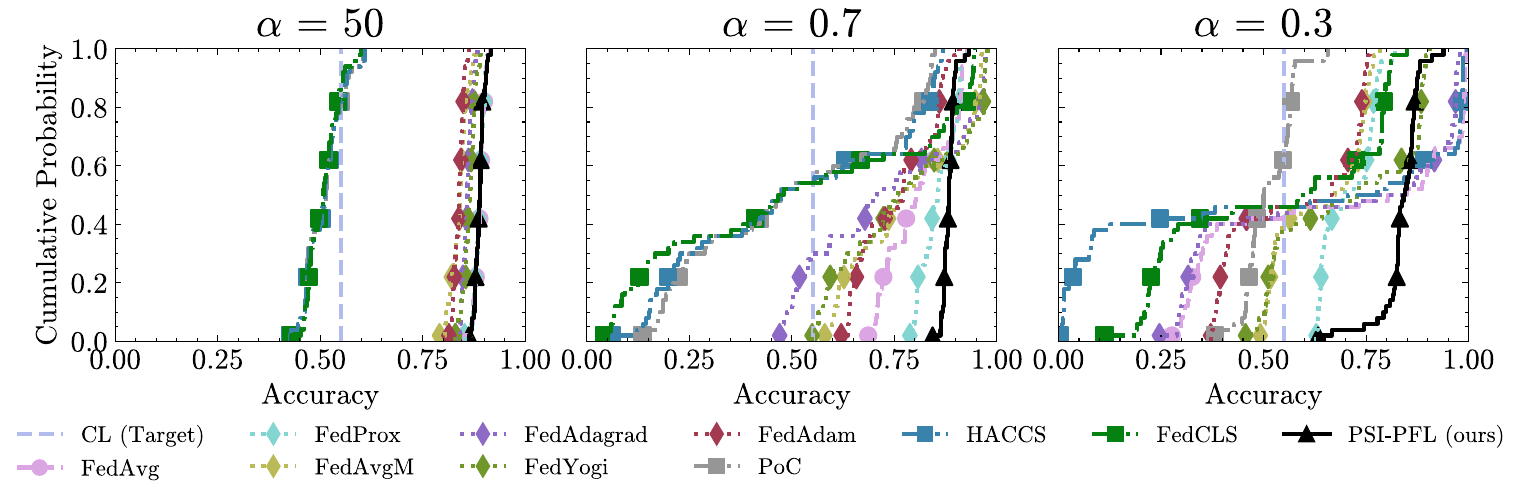}
        \caption{50 Clients}
        \label{fig:celeba_local_acc_dirichlet_50_CLI_LS_all_Alp}
    \end{subfigure}
    
    \vspace{0.5cm} 
    
    \begin{subfigure}{\textwidth}
        \centering
        \includegraphics[width=1\textwidth]{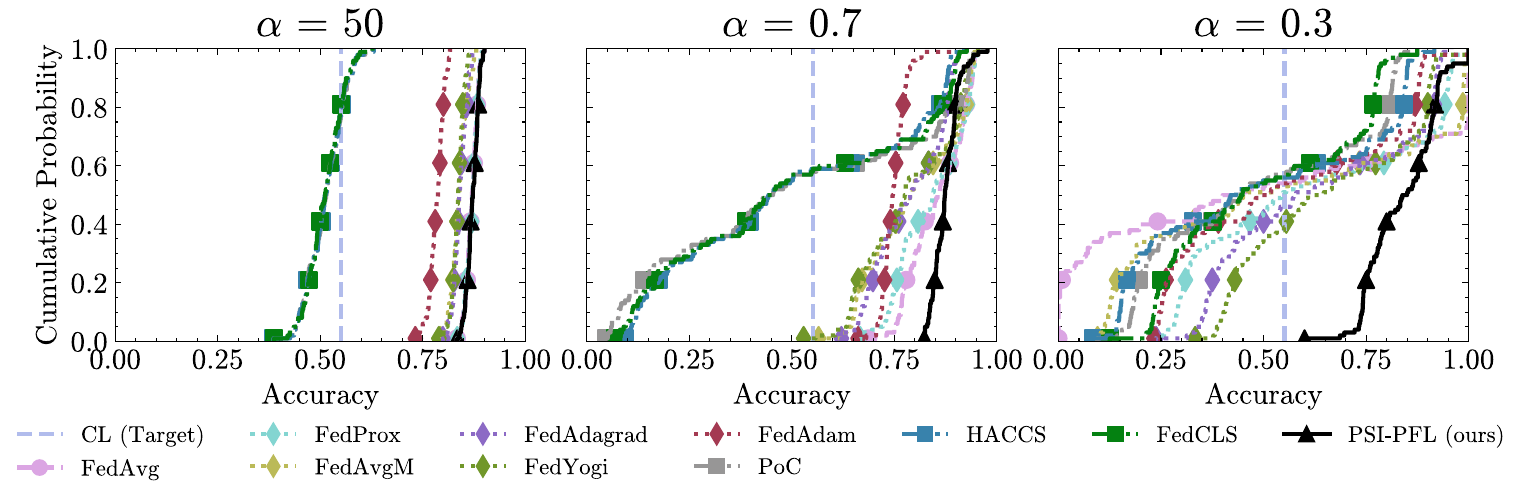}
        \caption{100 Clients}
        \label{fig:celeba_local_acc_dirichlet_100_CLI_LS_all_Alp}
    \end{subfigure}

    \caption{Local test accuracy ECDF for all baselines for the CelebA dataset.}
    \label{fig:celeba_three_images_vertical}
\end{figure}


\begin{figure}[ht]
\centering
\includegraphics[width=1\textwidth]{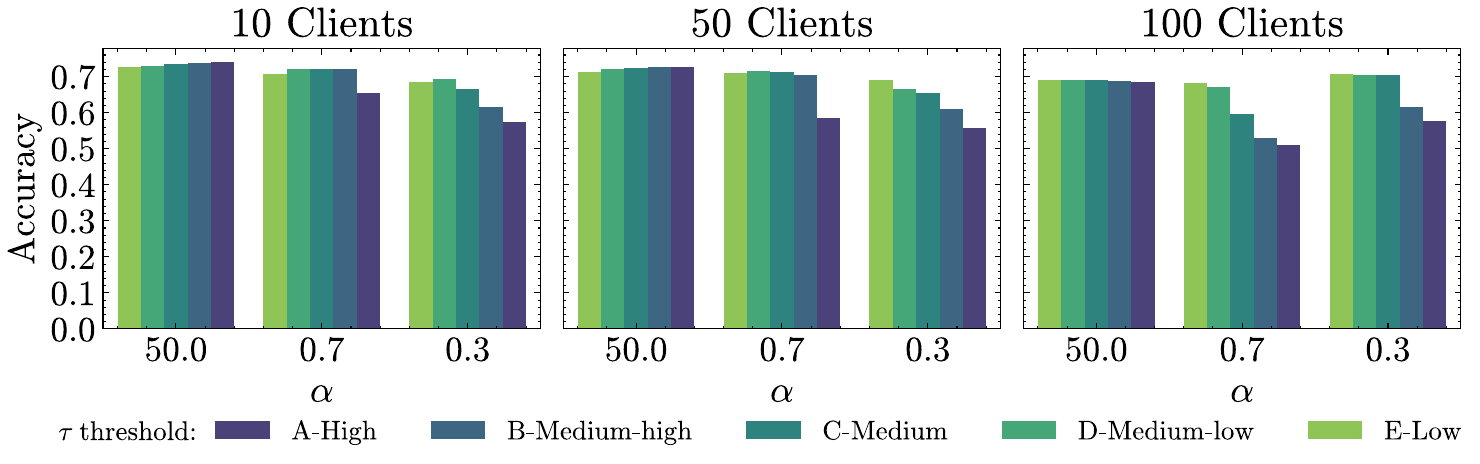}
\caption{Global test accuracy for PSI-PFL thresholds ($\tau$) for the Sent140 dataset.}
\label{fig:sent140_psi_thr_acc_dirichlet_all_CLI_LS_all_Alp}
\end{figure}

\begin{figure}[ht]
\centering
\includegraphics[width=1\textwidth]{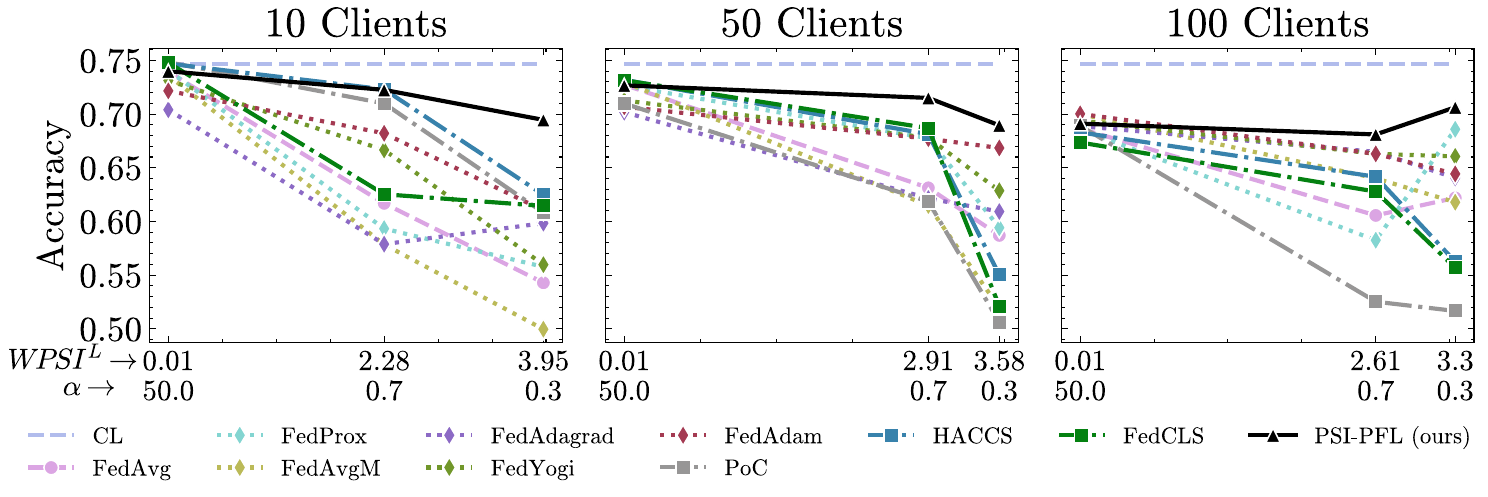}
\caption{Global test accuracy for all baselines and Sent140 dataset.}
\label{fig:sent140_global_acc_dirichlet_all_CLI_LS_all_Alp}
\end{figure}

\begin{figure}[htp]
    \centering
    \begin{subfigure}{\textwidth}
        \centering
        \includegraphics[width=1\textwidth]{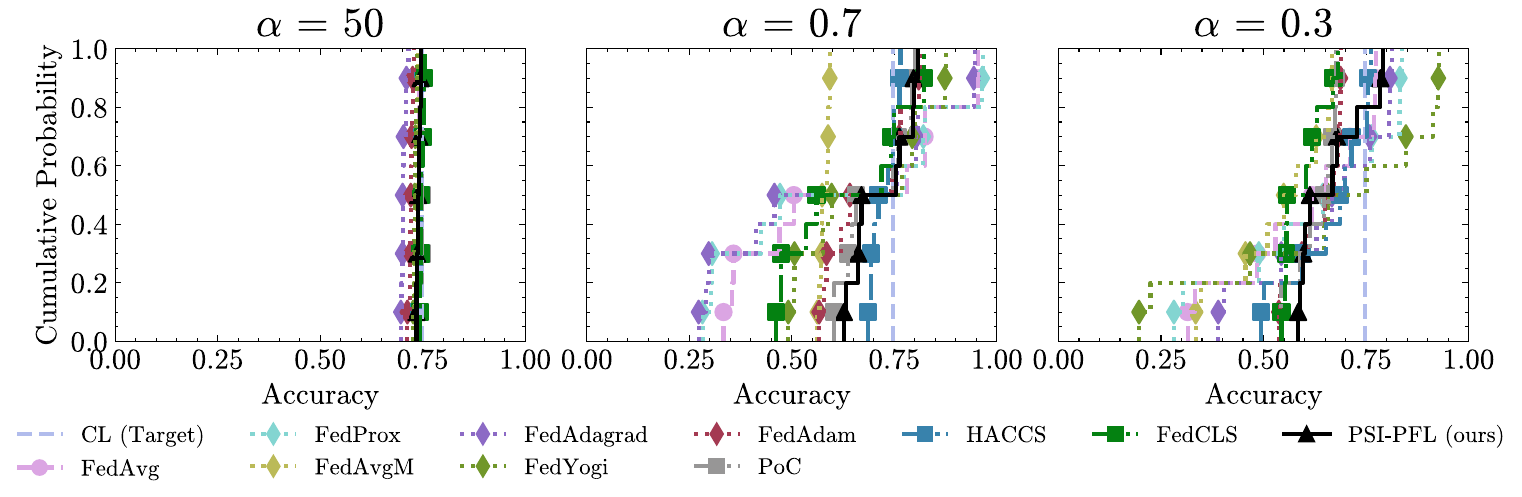}
        \caption{10 Clients}
        \label{fig:sent140_local_acc_dirichlet_10_CLI_LS_all_Alp}
    \end{subfigure}
    
    \vspace{0.5cm} 
    
    \begin{subfigure}{\textwidth}
        \centering
        \includegraphics[width=1\textwidth]{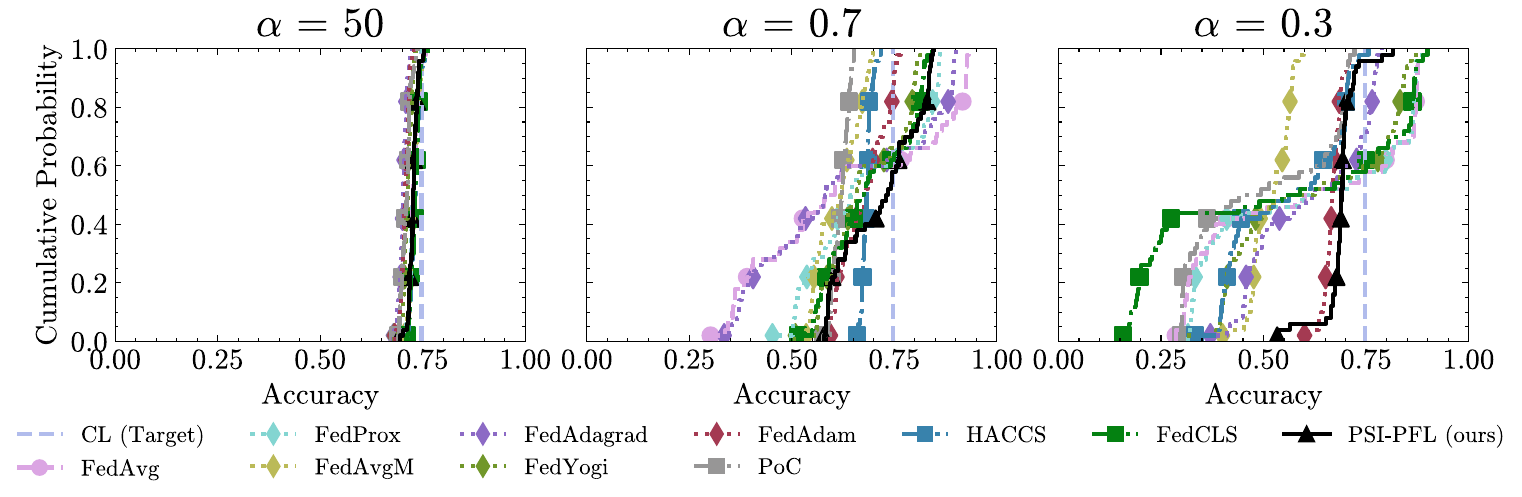}
        \caption{50 Clients}
        \label{fig:sent140_local_acc_dirichlet_50_CLI_LS_all_Alp}
    \end{subfigure}
    
    \vspace{0.5cm} 
    
    \begin{subfigure}{\textwidth}
        \centering
        \includegraphics[width=1\textwidth]{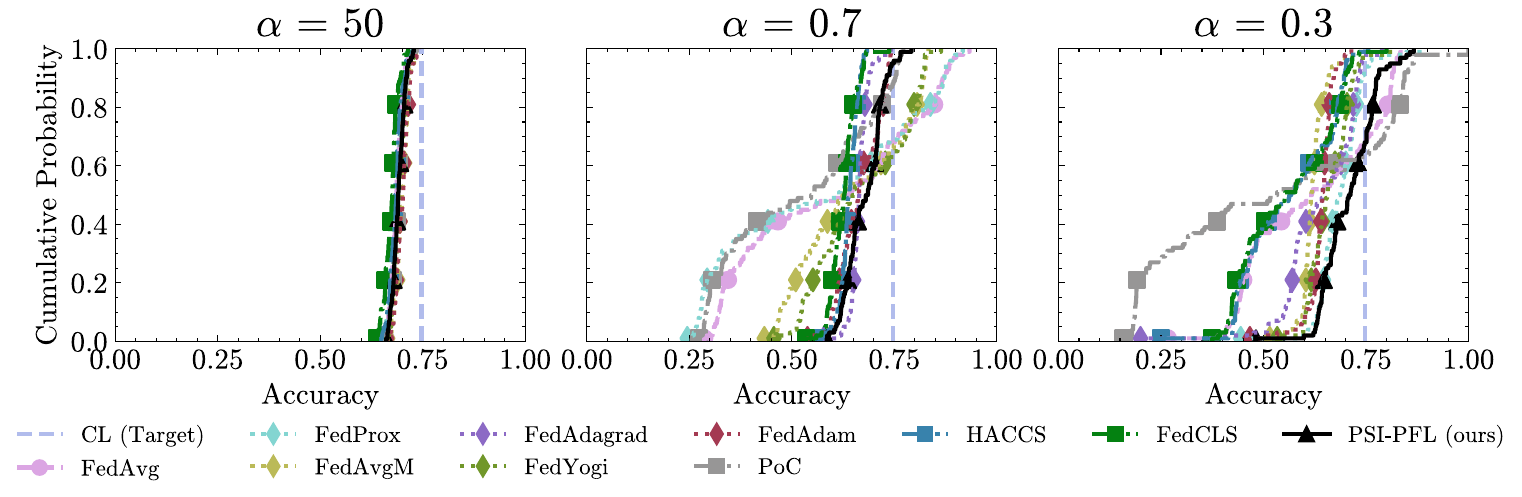}
        \caption{100 Clients}
        \label{fig:sent140_local_acc_dirichlet_100_CLI_LS_all_Alp}
    \end{subfigure}

    \caption{Local test accuracy ECDF for all baselines for the Sent140 dataset.}
    \label{fig:sent140_three_images_vertical}
\end{figure}


\end{document}